\documentclass{article}

\usepackage[preprint]{neurips_2025}

\usepackage[utf8]{inputenc} % allow utf-8 input
\usepackage[T1]{fontenc}    % use 8-bit T1 fonts
\usepackage[colorlinks]{hyperref}       % hyperlinks
\usepackage{url}            % simple URL typesetting
\usepackage{booktabs}       % professional-quality tables
\usepackage{amsfonts}       % blackboard math symbols
\usepackage{nicefrac}       % compact symbols for 1/2, etc.
\usepackage{microtype}      % microtypography
\usepackage{xcolor}         % colors

\usepackage{wrapfig}
\usepackage{multirow}
\usepackage{algorithm}%
\usepackage{algorithmicx}%
\usepackage{algpseudocode}%
\usepackage{amsmath,amssymb}%
\usepackage{amsthm}%
\usepackage[capitalise]{cleveref}
\usepackage{orcidlink}
\usepackage{subcaption}
\usepackage{enumitem}

\newtheorem{theorem}{Theorem}%  meant for continuous numbers
\newtheorem{corollary}{Corollary}% 
\newtheorem{definition}{Definition}%

\title{
Vine Copulas as Differentiable Computational Graphs
}

\author{%
  Tuoyuan~Cheng\orcidlink{0000-0001-5170-6546}
    \\
  Risk Management Institute\\
  National University of Singapore\\
  \texttt{tuoyuan.cheng@nus.edu.sg}\\
  \And
  Thibault~Vatter\orcidlink{0000-0001-9212-0218}\\
  University of Applied Sciences Western Switzerland (HES-SO)\\
  \texttt{thibault.vatter@hesge.ch}\\
  \AND
  Thomas~Nagler\orcidlink{0000-0003-1855-0046}\\
  LMU Munich and the Munich Center for Machine Learning\\
  \texttt{mail@tnagler.com}
  \AND
  Kan~Chen\orcidlink{0000-0002-1760-0876}\\
  Risk Management Institute\\
  Department of Mathematics\\
  National University of Singapore \\
  Singapore, Singapore \\
  \texttt{kan.chen@nus.edu.sg} \\
}

\begin{document}

\maketitle

\begin{abstract}
	% The abstract paragraph should be indented \nicefrac{1}{2}~inch (3~picas) on both the left- and right-hand margins. Use 10~point type, with a vertical spacing (leading) of 11~points.  The word \textbf{Abstract} must be centered, bold, and in point size 12. Two line spaces precede the abstract. The abstract must be limited to one paragraph.
	% Capturing complex multivariate dependencies for conditional sampling and quantile estimation in high dimensions remains a challenge. 
	% Vine copulas offer flexibility by decomposing joint distributions into cascades of more manageable bivariate copulas, yet their sequential (inverse) Rosenblatt transforms have lacked a unified, scalable computational framework.
    Vine copulas are sophisticated models for multivariate distributions and are increasingly used in machine learning. To facilitate their integration into modern ML pipelines, we introduce the vine computational graph, a DAG that abstracts the multilevel vine structure and associated computations. On this foundation, we devise new algorithms for conditional sampling, efficient sampling-order scheduling, and constructing vine structures for customized conditioning variables. We implement these ideas in $\verb|torchvinecopulib|$, a GPU-accelerated $\verb|Python|$ library built upon $\verb|PyTorch|$, delivering improved scalability for fitting, sampling, and density evaluation.
    Our experiments illustrate how gradient flowing through the vine can improve Vine Copula Autoencoders and that incorporating vines for uncertainty quantification in deep learning can outperform MC-dropout, deep ensembles, and Bayesian Neural Networks in sharpness,  calibration, and runtime. By recasting vine copula models as computational graphs, our work connects classical dependence modeling with modern deep-learning toolchains and facilitates the integration of state-of-the-art copula methods in modern machine learning pipelines.
\end{abstract}

\section{Introduction}\label{sec:introduction}
% All headings should be lower case (except for first word and proper nouns), flush left, and bold.
% \TYcom{Comment: 
% hook, context, gap, solution, contributions, roadmap\\
% }
% All headings should be lower case (except for first word and proper nouns), flush left, and bold.
% \TYcom{Comment: 
% hook, context, gap, solution, contributions, roadmap\\
% }
Vine copulas are graphical models that provide a flexible framework for modeling complex multivariate distributions \citep{czado_analyzing_2019,czado_vine_2022}. By decomposing a $d$-dimensional joint distribution into its one-dimensional margins and a cascade of bivariate copulas, vine copula models allow for capturing various marginal behaviors (skewness, light/heavy tails) and dependence asymmetries.

Vine copulas have also gained traction in the machine learning community, particularly in the context of probabilistic modeling and uncertainty quantification.
Recent applications include variational inference \citep{Tran2015}, causal inference \citep{Lopez-Paz2016,Tagasovska2019a}, clustering \citep{tekumalla2017vine, sahin2022vine}, regression \citep{kraus_d_vine_2017,tepegjozova_nonparametric_2022,Nagler2018}, Bayesian optimization \citep{parkbotied}, privacy-preserving inference \cite{Gambs2021}, retrospective uncertainty \citep{tagasovska2023retrospective}, and conformal prediction \citep{park2025semiparametric}. 
Many of those applications rely on the ability to (conditionally) sample from a fitted vine copula model, which is typically achieved through the Rosenblatt transform \citep{rosenblatt1952}, an idea similar to normalizing flows \citep{papamakarios2021normalizing}. This involves complex algorithms and substantial computational effort, especially  for large data sets. While efficient algorithms and software implementations exist in various programming languages \citep{vinecopula2024, VineCopulaMatlab, rvinecopulib, pyvinecopulib}, they currently do not exploit advances in GPU computing and therefore do not integrate well with modern ML pipelines. 

% Concurrently, computational graphs, typically \emph{directed acyclic graphs} (DAGs), have become the backbone of modern deep learning, providing modular model construction, automatic differentiation, and seamless GPU acceleration \citep{paszke_pytorch_2019}.

\paragraph{Main contributions} 

Our work bridges the gap between vine copulas and computational graphs, increasing the power and accessibility of vine copulas in modern machine learning applications.
\begin{enumerate}[leftmargin=16pt]
	\item We introduce the \emph{vine computational graph} (VCG), which abstracts vine structures and associated computations into a DAG. This allows us to leverage the power of computational graphs for efficient sampling and inference in vine copula models and facilitates GPU acceleration.
	\item  Based on the VCG, we make several algorithmic advances:
\begin{enumerate}[label=(\roman*)]
	\item Efficient conditional sampling through graph traversal on the VCG.
	\item Sampling order scheduling and source vertices selection algorithm that minimizes computational load in sampling and quantile computations.
	\item Vine graph construction admitting pre-specified sets of conditioning variables.
\end{enumerate}
\item The VCG and algorithms are implemented in \texttt{torchvinecopulib}, a GPU-accelerated \texttt{PyTorch} library designed to deliver improved scalability for fitting, sampling, and density evaluation.
\end{enumerate}

By recasting vine copula models as computational graphs, our work connects classical dependence modeling with modern deep-learning toolchains and facilitates the integration of state-of-the-art copula methods in \texttt{PyTorch} pipelines.

% \paragraph{Outline}
% % 
% % The remainder of the paper is organized as follows.
% \cref{sec:vcg} formalizes the vine computational graph concept.
% \cref{sec:sampling} presents our graph-traversal-based sampling algorithm.
% \cref{sec:sample_order} describes the sampling order scheduling algorithm.
% \cref{sec:rvine:construct} details vine construction algorithm.
% In \cref{sec:torchvine}, we introduce \texttt{torchvinecopulib}, our \texttt{PyTorch} implementation with GPU acceleration.
% \cref{sec:pred_intvl} demonstrates how vines retrospectively quantify the prediction intervals for trained neural networks, and \cref{sec:conclusion} concludes with future directions.

% For readers less familiar with vine copulas, \cref{sec:pcc} provides a three-dimensional example where an R-vine coincides with both a C-vine and a D-vine.
% Comprehensive surveys on vine copula can be found in \cite{joe_dependence_2014,czado_analyzing_2019,czado_vine_2022}.
% 
% 
% 

\section{Background on vine copula models}\label{sec:literature}
% All headings should be lower case (except for first word and proper nouns), flush left, and bold.
% \TYcom{Comment: definition (appendix), description (vertex/edge), justification (relationship to existing)\\}
% \TYcom{Comment: DAG definition is separate from highlighting (sampling order, detailed later);\\}

\paragraph{Copulas}
Copulas allow for flexible construction of a joint distribution with arbitrary one-dimensional margins \citep[see e.g., ][for textbook treatments]{nelsen_introduction_2006,joe_dependence_2014}.
Mathematically, a copula $C$ is a multivariate distribution with standard uniform margins.
The main justification for copula-based modeling is a theorem by Sklar \citep{sklar_fonctions_1959}:
if $X \in \mathbb{R}^d$ is a continuous random vector with joint density $f$ and  marginal densities $f_{0}, \dots, f_{d - 1}$, there exists a \emph{copula density} density $c$ such that
% \begin{align} \label{eq:sklar}
% F\left(x_0,\,\dots,\,x_{d-1} \right) &= C\left\{ F_{1}(x_1),\,\dots,\,F_{d-1}(x_{d-1}) \right\},\, \forall x \in \mathbb{R}^d.
% \end{align}
% And $C$ is unique if the margins are continuous. When it exists, the corresponding joint density is
\begin{align} \label{eq:density}
f(x_0, \ldots, x_{d-1}) = c\left\{ F_0(x_0), \ldots, F_{d-1}(x_{d-1}) \right\} \prod^{d-1}_{i=0} f_i(x_i).
\end{align}
The copula density $c$ is a regular joint density with uniform marginals. 
Accordingly, the joint log-likelihood is the sum of the marginal log-likelihoods and that of the copula. 
This can be exploited in a two-step procedure by first estimating each of the margins and then the copula \citep{genest1995semiparametric, joe1996estimation}.

\paragraph{Vines} While standard copulas are well understood and have found many applications, they are limited in their ability to model complex data structures.
This is because, in most copulas, dependencies between all subsets of variables are described by the same parametrization, which is often too restrictive.
Popularized in \citep{aas_pair-copula_2009}, vine copulas allow for a finer-grained modeling approach.
Following the seminal work of \citep{joe_families_1996} and \citep{bedford_vines_new_2002}, any copula density $c$ can be decomposed into a product of $d(d-1)/2$ bivariate (conditional) copula densities.
The order of conditioning in this decomposition can be organized using a graphical structure, called \emph{regular vine (R-vine)}---a sequence of trees $(\mathcal{V}_k, \mathcal{E}_k), k = 0, \dots, d-2$, satisfying the following conditions:
	\begin{enumerate}[leftmargin=18pt]
		\item $(\mathcal{V}_0, \mathcal{E}_0)$ is a spanning tree with nodes $\mathcal{V}_0=\{0, \dots, d-1\}$ and edges $\mathcal E_0$.
		\item For $k \ge 1$, $(\mathcal{V}_k, \mathcal{E}_k)$ is a spanning tree with nodes $\mathcal{V}_k=\mathcal{E}_{k-1}$ and edges $\mathcal{E}_k$.
		\item (\emph{Proximity condition}) When two nodes in $(\mathcal{V}_{k+1}, \mathcal{E}_{k+1})$ are joined by an edge, the corresponding edges in $(\mathcal{V}_k, \mathcal{E}_k)$ must share a common node.
	\end{enumerate}   
% \tn{we should probably add an example figure of a vine tree if space allows}
A \emph{vine copula} identifies each edge of the vine structure with a label $\{l, r | \mathcal{S}\}$ and a bivariate copula $C_{l, r|\mathcal{S}}$, called a \emph{pair-copula}.
The sets $\{l, r\} \in \mathcal{V}_0$ and $\mathcal{S} \subseteq \mathcal{V}_0 \setminus \{l, r\}$ are called the \emph{conditioned set} and \emph{conditioning set}, respectively, and we refer to \citep{czado_vine_2022} as well as \cref{def:vine_mst,def:vinecop_mst} in \cref{sec:vine:cop:def} for a precise definition.
The corresponding copula density then factorizes as follows:
\begin{align*}
c(u) = \prod_{k=0}^{d-2} \prod_{i=0}^{d-2-k} c_{l_i, r_i| \mathcal{S}_i} \bigl(C_{l_i|\mathcal{S}_i}(u_{l_i}|u_{\mathcal{S}_i}), C_{r_i|\mathcal{S}_i}(u_{r_i}|u_{\mathcal{S}_i}) \bigr),
\end{align*}
where $u_{\mathcal{S}} :=(u_j)_{j \in \mathcal{S}}$ is a subvector of $u =(u_0, \dots, u_{d-1}) \in [0,1]^d$ and $C_{j|\mathcal{S}}$ is the conditional distribution of $U_{j}$ given $U_{\mathcal{S}} = u_{\mathcal{S}}$.

\paragraph{Recursive computation with h-functions}
Regular vine graphs guarantee that the conditional distributions $C_{l_i|\mathcal{S}_i}$ can be computed recursively from only the pair-copula densities $c_{l_j,r_j|\mathcal{S}_j}$ appearing in the graph. Specifically, define the so-called $h$-functions
% \begin{align}
    $h_{l|r,\mathcal{S}}(u | v) = \int_{0}^u c_{l, r | \mathcal{S}}(s, v) ds$, and $h_{r|l,\mathcal{S}}(u | v) = \int_{0}^v c_{l, r | \mathcal{S}}(u, s) ds.$
% \end{align}
Now $C_{l | \mathcal{S}}$ can be expressed as
\begin{align} \label{eq:h-recursion}
  C_{l | \mathcal{S}}(u_{l} \mid u_{\mathcal{S}}) =   h_{l|r,\mathcal{S}'}(C_{l | \mathcal{S}'}(u_{l} \mid  u_{\mathcal{S}'}) \mid C_{r | \mathcal{S}'}(u_{r} \mid  u_{\mathcal{S}'})),
\end{align}
for some index $r \in \mathcal{S}$ and with reduced conditioning set $\mathcal{S}' = \mathcal{S} \setminus \{r\}$,
and similarly for $C_{r | \mathcal{S}}$. The identity \eqref{eq:h-recursion} can now be applied iteratively until the conditioning set is empty.

\paragraph{Conditional sampling}
The inverse Rosenblatt transform is commonly used to sample from a vine copula model \citep{czado_analyzing_2019}. It involves the inverse of the conditional distributions $C_{l_i | \mathcal{S}_i}$ and a permutation of the variables called \emph{sampling order}, denoted as
$\mathcal{S}_{\text{order}} = (l_0, l_1, \dots, l_{d-1})$ where $l_i \in \mathcal{V}_0$.
%for $0 \le i \le d-1$.
\begin{enumerate}[leftmargin=16pt]
    \item Simulate $(V_0, \dots, V_{d-1}) \sim \mathrm{Uniform}[0, 1]^d$.
    \item $U_{l_{d-1}} = V_{l_{d-1}},  U_{l_{d-2}} = C_{l_{d-2} | l_{d-1}}^{-1}(V_{l_{d-2}} | U_{l_{d-1}}),\ldots, U_{l_{0}}= C_{l_{0} | l_1,\dots, l_{d-1}}^{-1}(V_{l_{0}} | U_{l_1}, \dots, U_{l_{d-1}})$.
    \item Transform $X_0 = F_0^{-1}(U_0)$, \dots, $X_{d-1} = F_{d-1}^{-1}(U_{d-1})$.
\end{enumerate}
Similar to \eqref{eq:h-recursion}, the inverse conditional distributions $C_{l_i | \mathcal{S}_i}^{-1}$ with $\mathcal{S}_i = \{l_0, \dots, l_{i-1}\}$ 
 can be computed recursively from h-functions and their inverses.

\paragraph{Computational challenges}
There are $O(d^2)$ pair-copulas, and an h-function recursion starting from a pair-copula in the $k$-th tree level takes $\Omega(k)$ steps, although many intermediate results can be shared across pair-copulas. These recursions dominate the computational demand of vine copula models, and carrying them out efficiently is essential to keep runtimes manageable. How many different h-functions must be computed depends on the vine structure and sampling order. The following section presents new algorithms for selecting efficient structures and sampling orders.

\section{New computational model and algorithms}
% All headings should be lower case (except for first word and proper nouns), flush left, and bold.
% \TYcom{Comment: definition (appendix), description (vertex/edge), justification (relationship to existing)\\}
% \TYcom{Comment: DAG definition is separate from highlighting (sampling order, detailed later);\\}

We now introduce the \emph{vine computational graph} (VCG), a DAG that encodes a vine copula model's hierarchical structure and computational dependencies, and associated algorithms. The VCG makes the sequence of bivariate copula evaluations and $h$-function calls required for density evaluation, sampling, and inversion explicit. We opt to present only informal versions of the definitions and algorithms in this section to convey the core ideas more clearly to readers who may not be experts on vines. Complete, formal definitions and algorithms are provided in the appendix. 

\subsection{The vine computational graph}\label{sec:vcg}

\begin{definition}[informal]
The VCG is a DAG characterized by sets $\left\{ (\mathcal{V}_k, \mathcal{E}_k, \mathcal{E}_k^u, \mathcal{E}_k^d) \right\}_{k=0}^{d-2}$, where
\begin{itemize}[leftmargin=14pt]
  \item $\mathcal{V}_0 = \{0,1,\dots,d-1\}$ and
  $\mathcal{V}_k = \{\{l_i\mid \mathcal{S}_i\}\}_{i=0}^{d-k-1}$ are sets of \emph{variable (or data) nodes}, satisfying $|\mathcal{S}_i| = k$, $i\in\{0,1,\dots,d-1\}$, and $\mathcal{V}_k = \mathcal{E}_{k-1}$, for $k\ge 1$,
  \item $\mathcal{E}_k=\{\{l_i, r_i\mid \mathcal{S}_i\}\}_{i=0}^{d-k-2}$ are sets of \emph{copula nodes}, with elements representing bivariate copulas $C_{l_i, r_i|\mathcal{S}_i}$ for the dependence between the two variables $l_i$ and $r_i$ given $\mathcal{S}_i$, as in \cref{sec:literature},
  \item $\mathcal{E}_k^u = \{\{l_i \mid \mathcal{S}_i\}, \{r_i \mid \mathcal{S}_i\}\}_{i=0}^{2(d-k-2)}$ are sets of \emph{directed edges from upward}, with elements representing the data-flow dependencies from variable nodes to copula nodes, and
  \item $\mathcal{E}_k^d = \{\{l_i, r_i | \mathcal{S}_i\}, \{l_i \mid \mathcal{S}_i \cup \{r_i\}\}\}_{i=0}^{2(d-k-2)}$ are sets of \emph{directed edges to downward}, with elements representing the data-flow dependencies from copula nodes to variable nodes.
\end{itemize}
\end{definition}

Each copula node thus connects both upward and downward, maintaining the proximity and spanning-tree constraints of a vine.
Directed edges from variables to copulas encode fitting or conditioning operations, while edges from copulas to variables represent the application of $h$-functions. This organization mirrors the recursive nature of density calculations, as well as of the Rosenblatt transform and its inverse, thus enabling efficient traversal for both inference and simulation.
We refer to \cref{def:vine_dag} of \cref{sec:vine:cop:def} for a formal definition of the VCG.

% 
% For simplicity, we omit empty conditioning sets.
%
\begin{figure}[t!]
	\centering
	\begin{subfigure}[t]{0.495\textwidth}
		\centering
		\includegraphics[width=\linewidth]{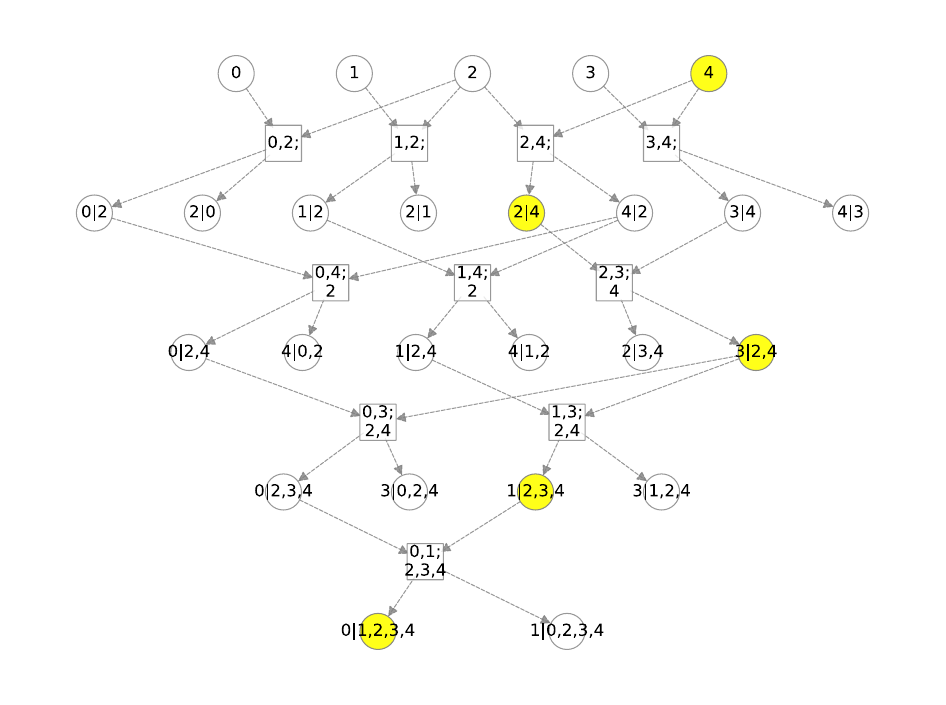}
		% \caption{Sampling order }
		\caption{}
		\label{fig:rvine_dag}
	\end{subfigure}
	\begin{subfigure}[t]{0.495\textwidth}
		\centering
		\includegraphics[width=\linewidth]{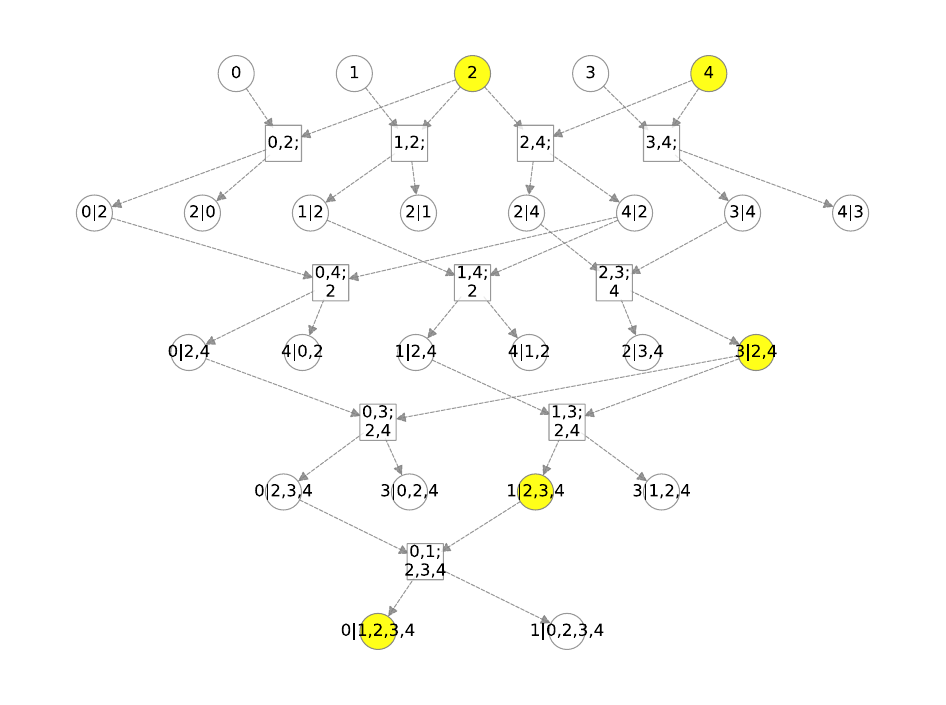}
		% \caption{Sampling order $(0,1,3)$}
		\caption{}
		\label{fig:rvine_dag_cond}
	\end{subfigure}
	\caption{
		A VCG with $d{=}5$ and (a) sampling with sampling order (right to left) $(0,1,3,2,4)$, (b) conditional sampling with sampling order $(0,1,3)$ and set of conditioning variables $\mathcal{S}_{cond}{=}\{2,4\}$.
	}
\end{figure}

\paragraph{Intepretation of the VCG} Illustrations of the VCG for several vine types are shown in \cref{fig:dvine_dag,fig:cvine_dag,fig:rvine_dag}.
Variable vertices in $\mathcal{V}_k$ at level $k$ are marked by a circle, with their singleton conditioned set followed by a vertical bar and then its (\emph{unordered}) conditioning set.
A bivariate copula vertex in $\mathcal{E}_k$ at level $k$, representing an attached bivariate copula, is marked by a square, with its (\emph{ordered}) bivariate conditioned set followed by a semicolon and then its (\emph{unordered}) conditioning set.
% 
% For $k\in\{0,1,\dots,d-2\}$, we have $|E_{k}^u| = |E_{k}^d| = 2|\mathcal{E}_k| = |\mathcal{V}_{k+1}|=2(d{-}k{-}1)$,
% where $(\mathcal{V}_k\cup\mathcal{E}_k, \mathcal{E}_k^u)$ forms a directed bipartite graph with all edges directed from vertices in $\mathcal{V}_k$ to vertices in $\mathcal{E}_k$,
% and $(\mathcal{E}_k\cup\mathcal{V}_{k+1}, \mathcal{E}_k^d)$ forms a directed bipartite graph with all edges directed from vertices in $\mathcal{E}_k$ to vertices in $\mathcal{V}_{k+1}$.

% A vine copula $\{(\mathcal{V}_k,\mathcal{E}_k,\mathcal{C}_k, \mathcal{E}_k^u, \mathcal{E}_k^d) | k\in\{0,1,\dots,d-2\}\}$ attaches to each bivariate copula vertex $e=\{l_i,r_i| \mathcal{S}_i\}\in\mathcal{E}_k$ a bivariate copula $C_{l_i,r_i| \mathcal{S}_i}\in \mathcal{C}_k$. 
Every bivariate copula vertex has an in-degree of $2$ from $\mathcal{E}_k^u$, signifying that the attached bivariate copula models the dependence between its two parent variables, and an out-degree of $2$ from $\mathcal{E}_k^d$, signifying the $h$-functions are applied to transform child conditional variables at the next level.
% Each bivariate copula vertex $l_i,r_i| \mathcal{S}_i$ has two parent variable vertices ($l_i| \mathcal{S}_i,~r_i| \mathcal{S}_i$) and two child variable vertices ($l_i| \mathcal{S}_i\cup\{r_i\},~r_i| \mathcal{S}_i\cup\{l_i\}$), where $l_i,r_i\in(\mathcal{S}\setminus\mathcal{S}_i),~l_i<r_i$ are elements in its conditioned set, $\mathcal{S}_i$ is its conditioning set, and $\mathcal{S}=\{0,\dots,d-1\}$ is the set of all indices of the $d$-dimensional unconditional copula.
% Edges directed from parent to bivariate copula indicate fitting the bivariate copula model onto the bivariate. 
% Edges directed from bivariate copula to child represent applying the $h$-function of the bivariate copula to transform the bivariate parent into child pseudo-observations at the next level.
% Edges directed from two parent variable vertices to a bivariate copula vertex indicate the bivariate copula is fitted using the parent, while edges directed from a bivariate copula vertex to two child variable vertices indicate applying the $h$-function of the bivariate copula to transform the parent into child pseudo-observations at the next level.
We may apply the $h$-function-inverse of the bivariate copula $\{l,r| \mathcal{S}\}$ onto $\{l| \mathcal{S}\cup\{r\}\}$ and $\{r| \mathcal{S}\}$ for the parent $\{l| \mathcal{S}\}$, or symmetrically onto $\{r| \mathcal{S}\cup\{l\}\}$ and $\{l| \mathcal{S}\}$ for the other parent $\{r| \mathcal{S}\}$.
The proximity condition ensures that two parent variable vertices can direct to the same bivariate copula vertex only if they (both the two parent variable vertices and the bivariate copula vertex) share the same conditioning set.
% 

% Certain vine structures impose additional constraints.
% D-vines as in \cref{fig:dvine_dag} requires all variable vertices at the top level to have an out-degree less than $3$, such that its spanning trees are of maximum diameter.
% C-vines as in \cref{fig:cvine_dag} places a central variable vertex at every level that directs to all bivariate copula vertices, such that its spanning trees are of minimum diameter.

% \TYcom{downward/upward visit and why directed edges\\}
We direct edges from variables $\to$ copulas $\to$ (child) variables, so that every $h$-function is abstracted as a \emph{downward visit} following the arrow, and every $h$-function-inverse is abstracted as an \emph{upward visit} against the arrow. Those directed edges locate variables and track pseudo-observation usage, while the ``visited" flags memoize intermediate results and facilitate reference-counting-based memory reclamation.
The resulting VCG provides the machinery for algorithms in the following sections.
% For brevity we will often write ``vine" when the context makes clear whether we refer to just the structure or the fully-fitted model.

\paragraph{Sampling orders} 
To sample from a vine copula, we must specify the order in which variables are generated. As described in \cref{sec:literature}, a \emph{sampling order} $\mathcal{S}_{\text{order}} = (l_0, l_1, \dots, l_{d-1})$ is a permutation of variables that determines the direction of traversal through the VCG.
This sequence prescribes that $l_{d-1}$ is sampled first, then $l_{d-2}$ conditioned on $l_{d-1}$, and so on, until $l_0$ is sampled conditioned on $l_1,\dots,l_{d-1}$.
In other words, $\mathcal{S}_{\text{order}}$ defines the set of \emph{source nodes} $\left\{ l_{i} ~\middle|~ \mathcal{S}_i\right\}$ with $\mathcal{S}_i = \{ l_{i+1}, \dots, l_{d-1} \}$ for $i=0,\dots,d-2$ and $\mathcal{S}_{d-1}=\emptyset$.
Each source node corresponds to a conditional variable whose conditioning set is determined by the remaining variables in $\mathcal{S}_{\text{order}}$ and all unsampled variables. Sampling proceeds from these source nodes upward through the VCG by applying inverse $h$-functions level by level.
For details, see \cref{sec:sampling,sec:sample_order}, as well \cref{def:sample_order} in \cref{sec:vine:cop:def}.

\subsection{Sampling via inverse Rosenblatt transform}\label{sec:sampling}
% All headings should be lower case (except for first word and proper nouns), flush left, and bold.
% \TYcom{Comment: description, correctness, justification (optimality or relationship to existing), time complexity\\}
% 
A sampling algorithm for VCGs is given in \cref{algo:sim-informal}.
Given a sampling order $\mathcal{S}_{\text{order}}$ of length $d$, we generate samples by traversing the vine computational graph (VCG) upward along $d$ distinct paths. Each path originates from a \emph{source vertex}, determined by the sampling order, 
%(see \cref{def:sample_order})
and ends at a corresponding \emph{target vertex} with the same conditioned variable but an empty conditioning set. 

\begin{algorithm}
	\caption{Unconditional sampling (informal)} \label{algo:sim-informal}
     \textbf{Input}: Vine computational graph and sampling order $ \mathcal{S}_\text{order}$.
   \begin{enumerate}[leftmargin=*]
    \item \textbf{Seeding:} For each $\{l \mid \mathcal{S}\}$, draw $u_{l \mid \mathcal{S}} \sim \mathcal{U}[0,1]$.
  \item \textbf{Recursive VCG ascent:} For $l \in \mathcal{S}_\text{order}$ do:
    \begin{enumerate}[label=2.\arabic*,leftmargin=*]
      \item Let $r$ be such that there is an edge from $\{l ,r \mid \mathcal S \setminus \{r\}\}$ pointing to $\{l \mid \mathcal S\}$ and compute  $u_{l | \mathcal S \setminus \{r\}} \leftarrow h^{-1}_{l \mid r, \mathcal{S}}(u_{l \mid \mathcal{S}} \mid u_{r \mid \mathcal{S}})$.
      If $u_{r \mid \mathcal{S}}$ is not yet available, apply the required $h$-function downward to generate it from its parents. Memoize the results.
     \item Update $\mathcal{S} \leftarrow \mathcal{S} \setminus \{r\}$ and continue with 2.1 as long as $\mathcal{S} \neq \emptyset$.
    \end{enumerate}
\end{enumerate}
\end{algorithm}

Using the VCG of \cref{fig:rvine_dag} as an example, the sampling order yields the sequences $\{4\}\!\to\!\{4\}$, $\{2|4\}\!\to\!\{2\}$, $\{3|2,4\}\!\to\!\{3|4\}\!\to\!\{3\}$, $\{1|2,3,4\}\!\to\!\{1|2,4\}\!\to\!\{1|2\}\to\{1\}$, and $\{0|1,2,3,4\}\to\{0|2,3,4\}\!\to\!\{0|2,4\}\!\to\!\{0|2\}\!\to\!\{0\}$.
Notice when ascending $\{1|2,4\}\!\to\!\{1|2\}$, the opposite parent $\{4|2\}$ is not yet visited, thus triggering a downward visit $\{4\}\!\to\!\{4|2\}$ via a $h$-function call.

This process guarantees that all pseudo-observations required by $h$-functions and their inverses are available when needed. The full algorithm requires $\mathcal{O}(d^2)$ $h$-function or inverse $h$-function evaluations, and we refer to \cref{sec:sample_order} for a detailed discussion of the complexity.
Full algorithmic details can be found in \cref{algo:sim} and its description in \cref{sec:algo}.

\paragraph{Conditional sampling}
The process generalizes seamlessly to conditional sampling when a subset of variables is held fixed, that is by seeding the VCG with observed values at a subset $\mathcal{S}_{\text{cond}} \subset \mathcal{V}_0$ of the target vertices with empty conditioning sets. 
For a given sampling order over all variables, any modification that ascends the shallowest $|\mathcal{S}_\text{cond}|$ source vertices onto the top level gives a feasible sampling order for conditional sampling.
Sequentially from the shallowest to the deepest, we take the remaining $|\mathcal{V}_0\setminus \mathcal{S}_\text{cond}|$ vertices as source vertices to ascend along their respective paths until reaching their target vertices at the top level.

% Given a bivariate copula vertex in $\mathcal{E}_k$, applying its $h$-function requires both parent variable vertices in $\mathcal{V}_k$ be visited, while applying its $h$-function-inverse requires one of its child variable vertices in $\mathcal{V}_{k+1}$ and the corresponding parent variable vertex in $\mathcal{V}_{k}$ with a different conditioned set (the ``opposite" parent) are visited. 
% In either case, after the operation, two parents and one child of the bivariate copula vertex are visited.
% 
% The vine computational graph's directed edges and ``visited" flags ensure no redundant calls and enable efficient memoization of intermediate pseudo–observations.
% The vine computational graph clearly details the sequential order of $h$-function-inverse, and tracks variable usages and $h$-function calls.
% The source vertices are arranged such that their paths by $h$-function-inverse must not intersect at any bivariate copula vertex. 
% Along each path, iteratively we use the $h$-function-inverse of a bivariate copula to walk from the deeper variable vertex to its corresponding upper variable vertex with the same conditioned set, where we may (recursively) request $h$-functions for the other upper (``opposite" parent) variable vertex if not yet visited.

As an illustration, consider the VCG of \cref{fig:rvine_dag_cond}. 
Using $\mathcal{S}_\text{order}=(0,1,3)$ as sampling order and $\mathcal{S}_\text{cond}=\{2,4\}$, the $|\mathcal{S}_\text{cond}|=2$ highlighted variable vertices with empty conditioning sets at the top level are given, while their conditioned sets are also subsets of the conditioning sets of the remaining $|\mathcal{V}_0\setminus\mathcal{S}_\text{cond}|=3$ deepest highlighted source vertices.
In this case, we are sampling $u_0,u_1,u_3$ conditioned on $u_2,u_4$.
% As a special case, the vine quantile regression has $|\mathcal{S}_{cond}|=d-1$.
% 
% 

\subsection{Sampling order scheduling} \label{sec:sample_order}
% All headings should be lower case (except for first word and proper nouns), flush left, and bold.
% \TYcom{Comment: description, correctness, justification (optimality or relationship to existing algo), complexity\\}

As discussed in \cref{sec:literature}, the sampling order plays an essential role in the computational demand of the sampling algorithm. No algorithm currently exists to exploit this fact.
We propose a new algorithm that greedily schedules sampling orders and source vertices that invoke the least number of $h$-function calls during (conditional) sampling.

\begin{algorithm}
	\caption{Sampling order scheduling (informal)} \label{algo:scheduling-informal}
     \textbf{Input}: Vine computational graph.
     
   \begin{enumerate}[leftmargin=*]
    \item Initialize $\mathcal{S}_\text{order} = ()$.
    \item For levels $k = d - 2, \dots, 0$:
    \begin{enumerate}[label=2.\arabic*,leftmargin=12pt]
    \item Pick the copula vertex in level $k$ that lies on the path prescribed by the current  $\mathcal{S}_\text{order}$.
    \item For the two potential next vertices (in level $k - 1$), query the number of required (inverse) $h$-function calls.
    \item Update $\mathcal{S}_\text{order} \leftarrow (\mathcal{S}_\text{order}, l_k)$, where $l_k$ is the vertex requiring fewer calls.
\end{enumerate}
\end{enumerate}
\end{algorithm}
For conditional sampling, similarly we begin the above process with a set of conditioning variables $\mathcal{S}_\text{cond}$, but prioritize those in $\mathcal{V}_0\setminus\mathcal{S}_\text{cond}$ in the selection step, and terminates the process early when $|\mathcal{S}_\text{order}| = |\mathcal{V}_0 \setminus \mathcal{S}_\text{cond}|$. The formal algorithm is given in \cref{algo:source}.

% The $d$ source vertices are arranged such that their paths by $h$-function-inverse do not intersect at any bivariate copula vertex. 

\paragraph{Illustration}
To illustrate the algorithm, consider the $d{=}5$-dimensional vine in \cref{fig:rvine_dag}.
% Initially, we highlight all variable vertices at the top level. 
Enumerating elements from the conditioned set of the only bivariate copula vertex at the bottom level, we query the number of $h$-function calls for the two sampling orders separately $\mathcal{S}_\text{order}$ (\cref{def:sample_order}) if we use one to sample from the constructed vine using \cref{algo:sim}.
With $\mathcal{S}_\text{order}=(0)$, source vertices are $\{0|1,2,3,4\},\{1\},\{2\},\{3\},\{4\}$, which incur $7$ $h$-function calls for conditional sampling, visiting the nodes $\{1|2,3,4\}, \{1|2,4\}, \{3|2,4\}, \{1|2\}, \{4|2\}, \{2|4\},\{3|4\}$.
With $\mathcal{S}_\text{order}=(1)$, source vertices are $\{1|0,2,3,4\}, \{0\}, \{2\}, \{3\}, \{4\}$, which also incurs $7$ $h$-function calls for conditional sampling.
We select $\mathcal{S}_\text{order}=(0)$, using the smaller index in the event of a tie.
Moving up one level, the bivariate copula vertex not yet used for $h$-function-inverse in the current sampling order is $\{1,3;2,4\}$.
From its two child variable vertices, the sampling order $\mathcal{S}_\text{order}=(0,1)$ incurs $4$ $h$-function calls for $\{3|2,4\}, \{2|4\}, \{3|4\}, \{4|2\}$, while the sampling order $\mathcal{S}_\text{order}=(0,3)$ incurs $5$ for $\{1|2,4\}, \{1|2\}, \{4|2\}, \{2|4\}, \{1|2,3,4\}$, so we select $\mathcal{S}_\text{order}=(0,1)$.
We repeat the above query/selection/level-up process until we reach the top level, append the last element in $\mathcal{S}\setminus\mathcal{S}_\text{order}$, and arrive at $\mathcal{S}_\text{order}=(0,1,3,2,4)$, which gives the source vertices $\{4\}$, $\{2|4\}, \{3|2,4\}, \{1|2,3,4\}, \{0|1,2,3,4\}$ that incur $1$ $h$-function call to visit $\{4|2\}$ during sampling.
% 
% \subsubsection{Discussion}
% \noindent
\paragraph{Discussion}

\begin{wrapfigure}{r}{0.551\textwidth}
% \begin{figure} 
\vspace{-20pt}
	\flushright
        \includegraphics[width=.551\textwidth]{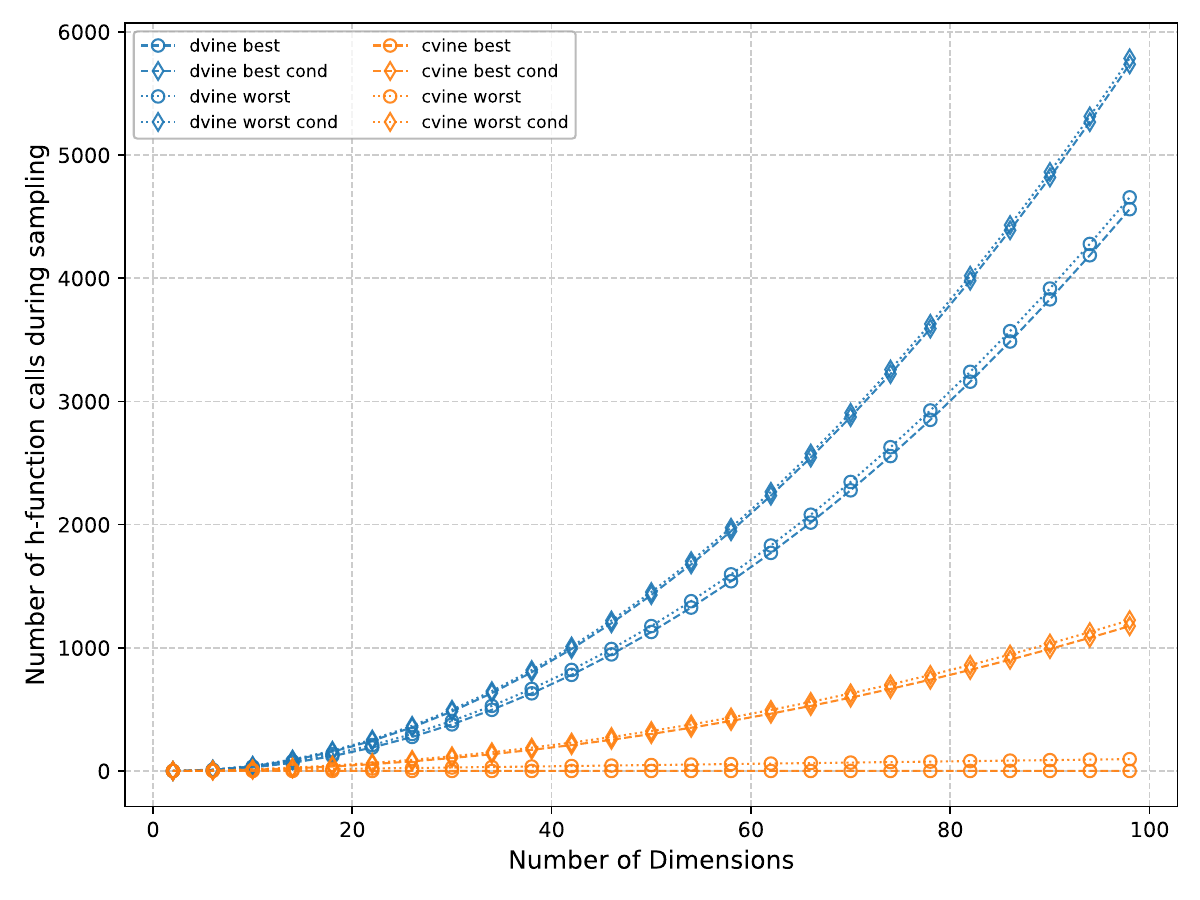}
	\caption{
		Number of $h$-function calls versus dimension $d$ for C-vines (orange) and D-vines (blue) under best (minimal) and worst (maximal) sampling orders. Circles show full sampling; diamonds show conditional sampling fixing half of the order. 
        % Counts are exact thus no error bar.
	}
	\label{fig:num_hfunc}
\vspace{-12pt}
% \end{figure}
\end{wrapfigure}
% An example is in \cref{fig:rvine_dag_cond}.
% 

The algorithm is correct as the resulting sampling order is in the solution set of an existing algorithm \citep{cooke_sampling_2015,zhu_common_2020}.
Given a $d$-dimensional vine, there are $2^{d-1}$ possible sampling orders for full sampling using inverse Rosenblatt transforms, each invoking a fixed number of $h$-function-inverse (\cref{sec:sampling}) but varying in the number of $h$-function-inverse calls, leading to varied sampling efficiencies.
Given a sampling order for full sampling, any modification that elevates the shallowest $|\mathcal{S}_\text{cond}|\in\{1,2,\dots,d-1\}$ source vertices onto the top level gives a feasible sampling order for conditional sampling.
Rather than querying exhaustively, our algorithm identifies the optimal one after $\mathcal{O}(d)$ queries, where each query incurs a VCG traversal of $\mathcal{O}(d^2)$, giving overall $\mathcal{O}(d^3)$ complexity.
The optimality proof in greedoid language is given in \cref{sec:sample_order:optimal}.

By greedily selecting the one incurring less/more $h$-function calls, we arrive at the best/worst sampling orders. We chart the relationship between the number of $h$-function calls during sampling and the number of dimensions $d$ for C-vines and D-vines in \cref{fig:num_hfunc}.
The dominant factor is the vine's structure itself.
% We further charted for conditional sampling when the sampling order is the first half of corresponding best/worst sampling orders for full sampling.
% The complexity is related to both structure and sampling order, with the former being much more dominant.
A C-vine (\cref{fig:cvine_dag}) has $\mathcal{O}(1)$ complexity with the best sampling order and $\mathcal{O}(d)$ complexity with the worst sampling order, while a D-vine (\cref{fig:dvine_dag}) has $\mathcal{O}(d^2)$ in both cases.
Given $d$-dimensional vine, conditional sampling has less $h$-function-inverse calls compared to full sampling but incurs more $h$-function calls.
Overall, C-vines achieve substantially higher sampling efficiency than D-vines in both full and conditional sampling.
\subsection{Vine graphs for pre-specified conditioning variables} \label{sec:rvine:construct}
% All headings should be lower case (except for first word and proper nouns), flush left, and bold.
% \TYcom{Comment: description, correctness, justification (optimality or relationship to existing algo), complexity\\}

To select the nested set of trees that form a vine structure, a popular approach is Dismann's algorithm \citep{dismann_selecting_2013}.
At a given level, it works by computing the absolute value of a dependence metric (e.g., Kendall's $\tau$ \citep{kendall_treatment_1945}) for all possible pairs of variable vertices allowed by the proximity condition, and then picking those forming a maximum spanning tree (MST).
% Several choices for the dependence metric $|f(\cdot,\cdot)|$ in $w$ are possible; e.g., Kendall's $\tau$ \citep{kendall_treatment_1945}, mutual information \citep{purkayastha_fastmi_2024}, Ferreira's tail dependence coefficient (TDC) \citep{ferreira_nonparametric_2013}, and Chatterjee's $\xi$ \citep{chatterjee_new_2021,lin_boosting_2023}.
% An adaptation of the general algorithm to select only C- or D-vines is discussed in  \cref{sec:dcvine}.
However, a vine graph restricts the choice of valid sampling orders. This poses a challenge for conditional sampling because a given vine might not allow closed-form sampling from an arbitrary conditioning set.
Solutions exist for C- and D-vines \citep{bevacqua2017cdvinecopulaconditional, aas2021explaining}, but, to the best of our knowledge, no algorithm exists in the literature for selecting R-vines with pre-specified conditioning sets.
In \cref{algo:kruskal-informal}, we propose a two-stage Kruskal's algorithm \citep{kruskal_shortest_1956} to fill this gap and refer to  \cref{algo:vine} in \cref{sec:algo} for the formal version.
\begin{algorithm}[h!]
	\caption{Vine selection for conditional sampling (informal)} \label{algo:kruskal-informal}
     \textbf{Input}: Set of conditioning variables $\mathcal{S}_{cond}$. 

 \begin{enumerate}[leftmargin=*]
     \item Initialize variable vertices in $\mathcal{V}_0$ with empty conditioning sets.
     \item For levels $k = 0, \dots, d - 2$:
   \begin{enumerate}[label = 2.\arabic*,leftmargin=*]
        \item Enumerate all pairs of variable vertices from $\mathcal{V}_k$ that share the same conditioning set.
        \item Append their child copula vertices to a candidates list $\mathcal{E}$ sorted according to the absolute value of the chosen dependence metric.
        \item Initialize $\mathcal{E}_{k}$ to the empty set and split the candidates list $\mathcal{E}$ into two parts: $\mathcal{E}_{s1}$, where $\mathcal{E}_{s1}$ contains edges whose conditioned set is a subset of $\mathcal{S}_{cond}$, and $\mathcal{E}_{s2} =\mathcal{E}\setminus\mathcal{E}_{s2}$.
        Then:
        \begin{enumerate}[label = 2.3.\arabic*, leftmargin=12pt]
            \item For each candidate edge in $\mathcal{E}_{s1}$: check whether adding it to $\mathcal{E}_{k}$ would create a cycle, add it if not, and stop if $|\mathcal{E}_{k}|=|\mathcal{S}_{cond}|-k-1$.
            \item For each candidate edge in $\mathcal{E}_{s2}$: check whether adding it to $\mathcal{E}_{k}$ would create a cycle, add it if not, and stop when $|\mathcal{E}_{k}|=d-k-1$.
        \end{enumerate}
        \item Fit bivariate copulas and compute pseudo-observations for the next level.
    \end{enumerate}
 \end{enumerate}
\end{algorithm}
Checking for cycles can be done efficiently using a union-find data structure \citep{cormen_introduction_2022} as in Kruskal's original MST algorithm \citep{kruskal_shortest_1956}.
It is easy to see that the two-stage structure does not incur additional complexity, so that step 2.3.1 of \cref{algo:kruskal-informal} is $\mathcal{O}(|\mathcal{E}|\log |\mathcal{E}|)$, where $|\mathcal{E}|$ is the number of edges in the candidates list $\mathcal{E}$.

Similarly to Dissman's algorithm, \cref{algo:kruskal-informal} is a heuristic with no formal optimality guarantees regarding the structure.
But this is not surprising, given that the space of vine structures grows super-exponentially in $d$ (i.e., too large for an exhaustive search) while offering too little structure for efficient exploration.
Note that \cref{algo:kruskal-informal} could also be used for regression context. Including all variables except the response in the conditioning set, one could extend the vine-based quantile regression approach of \citep{kraus_d_vine_2017, tepegjozova_nonparametric_2022, chang_prediction_2019} to R-vines.

\subsection{\texttt{torchvinecopulib}: a GPU-ready implementation} \label{sec:torchvine}
% All headings should be lower case (except for first word and proper nouns), flush left, and bold.
The VCG translates naturally into code: variable vertices become transient tensors of pseudo-observations, while bivariate-copula vertices are persistent model objects exposing primitive routines including h-function, h-function-inverse, distribution function, and density \citep{nagler_pyvinecopulib_2023,nagler_vinecopulib_2023}.
We built these ideas into \texttt{torchvinecopulib} (TVC) \citep{cheng_torchvinecopulib_2024}, a \texttt{Python} library built upon \texttt{PyTorch} and runs on CPUs or GPUs.
Given dimension $d$, the number of bivariate copulas is fixed at ${d}\choose{2}$, only their wiring in the VCG is learned.
Because the visit pattern during sampling is fully determined by the vine structure and the chosen source vertices, TVC releases intermediate tensors as soon as their reference count reaches zero, keeping peak memory roughly $\mathcal{O}(d)$.

\begin{figure}
    \vspace{-20pt}
	\centering
	\includegraphics[width=1\textwidth]{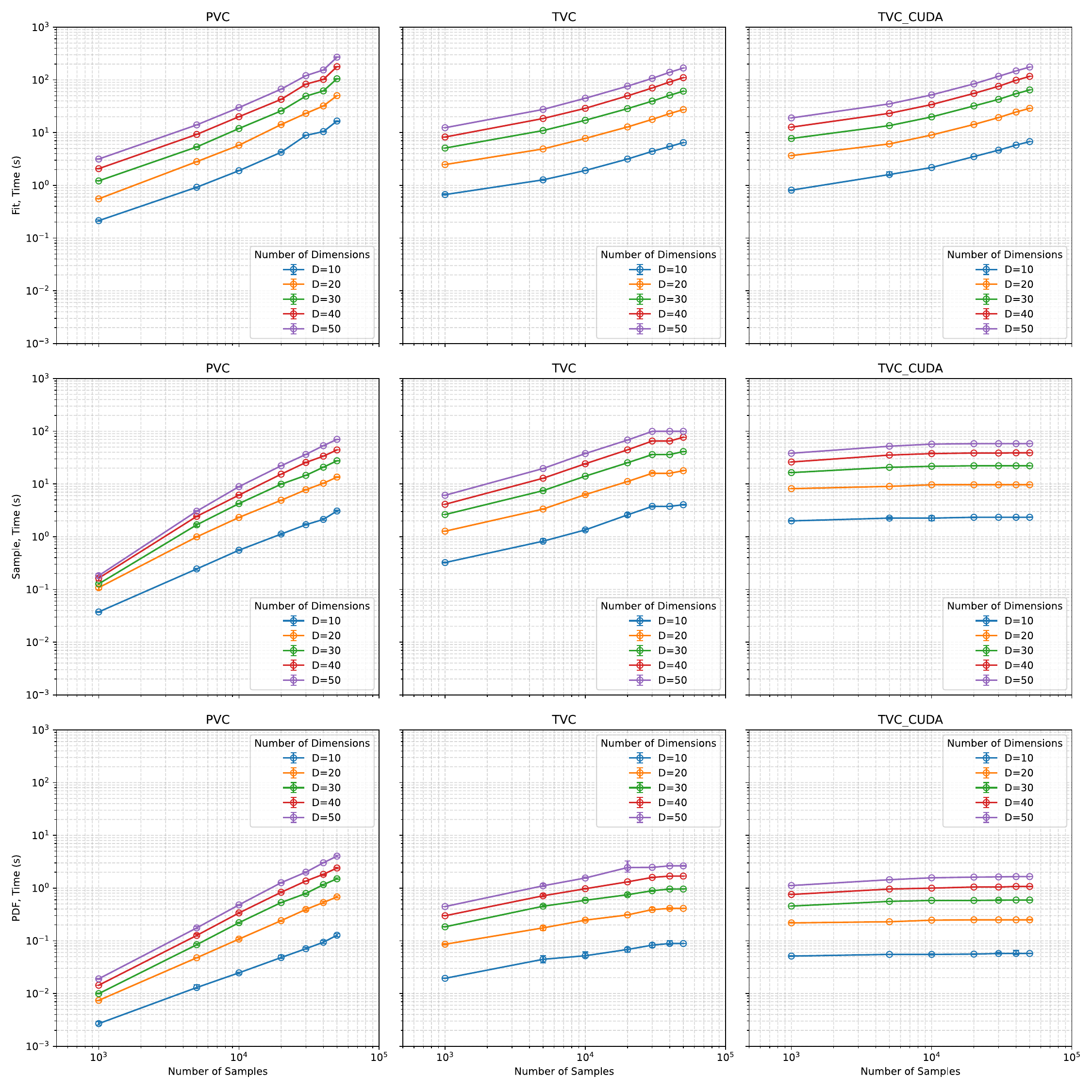}
	\caption{
	Runtimes for vine copula \emph{fit}, \emph{sample}, and \emph{density} tasks, using \texttt{pyvinecopulib}, \texttt{torchvinecopulib}-CPU/GPU.
	% Synthetic benchmarks span $d{=}10{-}50$ dimensions and $n{=}1{,}000{-}50{,}000$ observations. 
    Bars show the mean with $2.5\%-97.5\%$ quantiles on $100$ runs.
	}
	\label{fig:benchmark_time}
\end{figure}

% If a vine configuration (structure and bivariate dependence metric) captures most bivariate copula dependencies at shallower levels, leaving deeper bivariate copula independent, it enhances both fitting and sampling efficiency.
\cref{fig:benchmark_time} compares TVC with the \texttt{C++} backend \texttt{pyvinecopulib} (PVC).
For \emph{fit} task, PVC remains the fastest up to $n\approx20{,}000$, but TVC overtakes for larger sample size $n$.
For \emph{sampling} and \emph{density} tasks, TVC's vectorization delivers flatter scaling on CPUs and nearly constant time on GPUs.

TVC integrates with the \texttt{PyTorch} ecosystem---\texttt{autograd} support, batched tensor inputs, and GPU acceleration---making vine copulas a practical drop-in learning layer for a \texttt{PyTorch} pipeline.
% The computational graph decouples variable vertices and bivariate copula vertices, mirroring the separation of data and bivariate copula model objects in software implementations .
% Abstracting a vine into a DAG with highlighted source vertices reveals the sequential nature of both fitting and sampling.
% In our recent implementation using PyTorch for vectorization with GPU support \citep{cheng_torchvinecopulib_2024}, we treat data as transient streams and copula objects as persistent models encompassing a collection of functions including h-function, h-function-inverse, distribution function, and density.
% During vine fitting, we use data to fit bivariate copulas and visit variable vertices of deeper levels, in a feedforward manner.
% At level $k$ we fit bivariate copulas in $\mathcal{E}_k$, transform for data in $\mathcal{V}_{k+1}$, then release memory for all data in $\mathcal{V}_k$.
% Eventually, we retain fitted bivariate copulas in the vine model but discard data. 
% Due to proximity conditions, some variable vertices do not direct to a bivariate copula vertex, allowing the associated h-function calls to be skipped.
% During sampling, we walk h-function-inverse paths towards upper levels, returning only top-level data without altering the vine model.

\section{Experiments} \label{sec:experiments}

We revisit two recent applications to illustrate how neatly our VCG implementation and new algorithms integrate with modern deep learning pipelines.
Full ranges of hyperparameters and optimizer settings are listed in \cref{sec:expriment_details}, and our code repository provides details at \texttt{https://anonymous.4open.science/r/torchvinecopulib-7175/README.md}.
\subsection{Vine copula autoencoders} \label{sec:vcae}

Autoencoders (AEs, see e.g., \citep{hinton2006reducing,vincent2008extracting}) are a class of neural networks that learn to encode input data into a lower-dimensional representation and then decode it back to the original data. They are widely used for dimensionality reduction, feature extraction, and generative modeling.
In \citep{Tagasovska2018b}, the authors proposed the \emph{Vine Copula Autoencoder} (VCAE) to turn any AE into a generative model:
using vines to capture the dependencies between the latent variables, samples from the latent space are then decoded to the original space.
Note that training is not end-to-end, as the vine is trained separately from the AE.
Since vines can theoretically model any joint distribution, one advantage is that the AE can focus on minimizing the reconstruction error.
However, this separation of training can lead to suboptimal performance, as the AE may learn a representation that is ill-suited for vines, especially when the latent space is high-dimensional and complex.

We propose a novel joint training strategy VCAEs that enables the backpropagation of vine gradients through the AE. The goal is to fine-tune the latent representation to better align with the vine copula model, thereby improving the overall generative performance.
Our method proceeds in three stages:

\begin{enumerate}[leftmargin=*]
\item \textbf{Initial Training:} Train the AE to minimize the reconstruction error.
\item \textbf{Vine Fitting:} Fit a vine copula to the latent representation obtained from the training data.
\item \textbf{Fine-Tuning:} Retrain the AE by jointly minimizing the reconstruction error and the negative log-likelihood under the previously estimated vine copula. The vine parameters are kept fixed during this step, allowing gradients from the vine log-likelihood to propagate through the AE encoder.
\end{enumerate}

The first two steps are identical to the original VCAE approach, but the third step introduces a new loss term encouraging the AE to produce latent representations more consistent with the vine copula.

\begin{wraptable}{r}{0.5\textwidth}
\vspace{-6pt}
	\centering
	\caption{
		VCAE performances on MNIST. %Mean$\pm2\sigma$ over $100$ seeds.
		\\
		% Place one line space before the table title, one line space after the table title, and one line space after the table.
	}
	\label{tab:merged_results_mnist}
    \scriptsize
	\begin{tabular}{lccc}
		\toprule
		\textbf{Method} & \textbf{LogLik}  & \textbf{MMD} & \textbf{FID}  \\
		\midrule
		 Without refit (original) & $-44.7$          & $0.64$          & $6.65$          \\
    \textbf{With refit (ours)}     & $\mathbf{-28.4}$          & $\mathbf{0.50}$       & $\mathbf{6.42}$          \\
		\bottomrule
	\end{tabular}
\end{wraptable}

In \cref{tab:merged_results_mnist}, we compare the original VCAE with our new approach on the MNIST dataset \citep{lecun1998gradient}. 
We use a convolution-free AE composed of a fully connected encoder, mapping images to a 10-dimensional latent space with hidden layers of sizes 64 and 32. This symmetric decoder reconstructs the input from latent codes and is trained using the mean squared error loss. The results in \cref{tab:merged_results_mnist} show that our joint training strategy significantly improves performance, as measured by log-likelihood, maximum mean discrepancy (MMD, \citep{gretton2012kernel}), and Fréchet Inception Distance (FID, \citep{heusel2017gans}).

\subsection{Prediction intervals for neural networks via vine copulas} \label{sec:pred_intvl}
% All headings should be lower case (except for first word and proper nouns), flush left, and bold.
Modern neural networks (NNs) excel at point prediction, but typically do not provide uncertainty measures like \emph{prediction intervals} (PIs). Practitioners, therefore, turn to MC dropout \citep{kendall2017uncertainties}, deep ensembles \citep{lakshminarayanan2017simple}, or Bayesian NNs \citep{hernandez2015probabilistic}, each incurring a price, whether in terms of accuracy or runtime \citep{tagasovska2023retrospective}.
In contrast, the \emph{retrospective} vine approach simply ``plugs in" to any trained network: we collect each example's last-layer hidden features $\mathbf z\in\mathbb R^d$ and true response $y$, fit a nonparametric \citep{nagler2017nonparametric} $({d+1})$–dimensional vine copula on $(\mathbf z,y)$ (all $\mathbf z$ as $\mathcal{S}_{cond}$) (\cref{algo:vine}), choose the optimal vine sampling order (\cref{algo:source}), and then compute the $(2.5\%,97.5\%)$ PI via conditional quantiles $F^{-1}_{Y|\mathbf Z}(\frac{\alpha}{2}\mid\mathbf z)$ and $F^{-1}_{Y|\mathbf Z}(1-\frac{\alpha}{2}\mid\mathbf z)$ with $\alpha=0.05$ (\cref{algo:sim}). This lightweight procedure is task/architecture-agnostic and requires no NN retraining, adds little extra runtime, and guarantees no quantile-crossing due to the monotonicity of all h-functions.
\paragraph{Datasets and baselines}
We evaluate on two real-world regression tasks: California Housing \citep{pace1997sparse} and Online News Popularity \citep{fernandes2015proactive}. As baselines we compare against Deep Ensembles \citep{lakshminarayanan2017simple}, MC-dropout \citep{kendall2017uncertainties}, and Bayesian NNs \citep{hernandez2015probabilistic}, each tasked to produce $(2.5\%,97.5\%)$ PIs.
We evaluate on two real-world regression benchmarks: California Housing \citep{pace1997sparse} and Online News Popularity \citep{fernandes2015proactive}.
% As baselines we compare with Deep Ensembles \citep{lakshminarayanan2017simple}, MC-dropout \citep{kendall2017uncertainties}, and Bayesian NNs \citep{hernandez2015probabilistic}.
% 
\paragraph{Metrics}
We measure both \emph{sharpness} and \emph{calibration} via the mean \emph{interval (Winkler) score} (IS), which penalizes both interval width and miscoverage, as well as the mean \emph{pinball loss at low quantile} $2.5\%$ (PLL) and the mean \emph{pinball loss at high quantile} $97.5\%$ (PLH).
Over $100$ random train/val ($70\%/10\%$) splits and initialization seeds runs, we report the mean$\pm2\sigma$ variability of metrics on the fixed held test set ($20\%$) in \cref{tab:merged_results}.

\begin{table}[!ht]
\vspace{-6pt}
    \centering
    \caption{Performances of uncertainty quantification. Mean$\pm2\sigma$ over $100$ seeds.}
    \label{tab:merged_results}
    \scriptsize
    \begin{tabular}{l ccc ccc}
        \toprule
        \multirow{2}{*}{\textbf{Method}} 
        & \multicolumn{3}{c}{\textbf{California Housing}} 
        & \multicolumn{3}{c}{\textbf{Online News Popularity}} \\
        & \textbf{IS}      & \textbf{PLL}        & \textbf{PLH}        
        & \textbf{IS}      & \textbf{PLL}        & \textbf{PLH}       \\
        \midrule
        Ensemble   & $6.74 \pm 0.21$   & $0.07 \pm 0.01$   & $0.01 \pm 0.00$
                   & $7.08 \pm 0.83$   & $0.06 \pm 0.01$   & $0.12  \pm 0.02$  \\
        MC Dropout & $6.14 \pm 0.41$   & $0.05 \pm 0.01$   & $0.08 \pm 0.01$
                   & $7.14 \pm 0.90$    & $0.06 \pm 0.01$   & $0.12  \pm 0.02$  \\
        BNN        & $11.46   \pm 0.40$   & $0.19  \pm 0.01$   & $0.13   \pm 0.01$
                   & $10.86   \pm 1.54$   & $0.13  \pm 0.03$   & $0.14  \pm 0.02$  \\
        Vine       & $\mathbf{4.75  \pm 1.05}$   
                   & $\mathbf{0.04 \pm 0.01}$ 
                   & $\mathbf{0.079\pm 0.03}$ 
                   & $\mathbf{4.33  \pm 1.14}$ 
                   & $\mathbf{0.01\pm 0.00}$ 
                   & $\mathbf{0.10  \pm 0.03}$  \\
        \bottomrule
    \end{tabular}
\end{table}

% 
% 
% \begin{table}[!ht]
% \centering
% \caption{Predictive‐interval performance (mean\,$\pm2\sigma$ over 100 seeds).}
% \label{tab:merged_results}
% %
% \begin{subtable}{0.8\linewidth}
%   \centering
%   \caption{California Housing}
%   \begin{tabular}{lccc}
%     \toprule
%     \textbf{Method} & \textbf{IS} & \textbf{PLL} & \textbf{PLH} \\
%     \midrule
%     Ensemble   & $6.737\pm0.208$ & $0.069\pm0.004$ & $0.099\pm0.004$ \\
%     MC Dropout & $6.144\pm0.414$ & $0.046\pm0.005$ & $\mathbf{0.083\pm0.009}$ \\
%     BNN        & $11.462\pm0.397$& $0.191\pm0.008$ & $0.196\pm0.013$ \\
%     Vine       & $\mathbf{5.584\pm1.194}$ & $\mathbf{0.038\pm0.005}$ & $0.102\pm0.027$ \\
%     \bottomrule
%   \end{tabular}
% \end{subtable}

% \vspace{1em}

% \begin{subtable}{0.8\linewidth}
%   \centering
%   \caption{Online News Popularity}
%   \begin{tabular}{lccc}
%     \toprule
%     \textbf{Method} & \textbf{IS} & \textbf{PLL} & \textbf{PLH} \\
%     \midrule
%     Ensemble   & $7.081\pm0.832$  & $0.061\pm0.005$ & $0.116\pm0.016$ \\
%     MC Dropout & $7.138\pm0.896$  & $0.062\pm0.009$ & $0.117\pm0.015$ \\
%     BNN        & $10.852\pm1.540$ & $0.131\pm0.026$ & $0.141\pm0.018$ \\
%     Vine       & $\mathbf{4.706\pm0.722}$ & $\mathbf{0.008\pm0.001}$ & $\mathbf{0.110\pm0.018}$ \\
%     \bottomrule
%   \end{tabular}
% \end{subtable}
% \end{table}
% 
% 
\paragraph{Results}
Across both tasks, vines deliver the tightest and best-calibrated intervals by a clear margin.
% On California Housing, vines reduce the average IS to $4.75$,  lower than MC dropout ($6.14$) and Deep Ensembles ($6.74$), while achieving the smallest PLL. Its variance in IS over seeds is higher.
% On News Popularity, vines again yield the best average IS ($4.33$), better than the nearest competitor, and achieve the best calibration at the lower and upper tails.
By contrast, Deep Ensembles and MC-dropout perform comparably to each other, with moderately good intervals, but with substantially larger PLL. Our retrospective vine approach consistently outperforms in sharpness and calibration, without ever retraining the underlying network.
% 
% This lightweight, architecture-agnostic plug-in is very attractive given the growing environmental and financial costs of large-scale model training.

\section{Conclusion} \label{sec:conclusion}
% \TYcom{Comment: Question: Does the paper discuss the limitations of the work performed by the authors? (Checklist 2)\\}
% We presented the \emph{vine computational graph} (VCG), a DAG that abstracts both the multilevel structure of a vine with the sequence of h-function and h-function-inverse operations required for (conditional) sampling.
% By decoupling variable and copula vertices and orienting edges to reflect the data-flow dependencies, we reduced vine sampling into a graph-traversal problem.
% Building on this foundation, we devised algorithms to
% (i) perform conditional sampling including quantile regression, by implementing inverse Rosenblatt transforms as upward/downward visits
% (ii) select source vertices and sampling orders that minimize h-function invocations, and
% (iii) construct R-vine structures that admit specified set of conditioning variables via a two-stage Kruskal's MST.
% We packaged these ideas into \texttt{torchvinecopulib}, a \texttt{PyTorch} library with tensor support and GPU acceleration, achieving improved scalability.
% 
% In PI experiments on California Housing and Online News Popularity,
% our retrospective vines quantile regression approach outperformed MC-dropout, deep ensembles, and Bayesian NNs in sharpness and calibration (mean IS, PLL, and PLH), without retraining the original network or targeting specific quantiles during fitting.
% (including quantile regression) by plugging vine copulas into pre-trained neural networks without any retraining.
This work significantly advances the applicability of vine copula models in machine learning.
By translating vine copulas as computational graphs, we bridge the gap between classical dependence modeling and modern deep-learning toolchains. This opens the door to integrating uncertainty-quantification layers into any \texttt{PyTorch} pipeline, and suggests exciting directions for end-to-end training of copula layers, multi-stage vine architectures, and large-scale applications.
%in risk management and broader scientific domains.
% 
\paragraph{Limitations} The current implementation in \texttt{torchvinecopulib}  has not yet reached the state of maturity of other established libraries for vine copula modeling. In particular, it is limited regarding pair-copula families, the ability to model non-simplified vines \citep{nagler2025simplifiedvinecopulamodels}, marginal models, and user-friendliness. Another issue is the $\mathcal{O}(d^2)$ scaling of vine copula models in terms of memory and runtime, which can be overcome by optimized algorithms for truncated models.  
%We aim to address these issues in future work.
% \label{sec:limitations}
% \noindent
% First, like much of the existing vine copula literature, we assume that each conditional bivariate copula does not depend on the specific values of its conditioning variables \citep{nagler2025simplifiedvinecopulamodels}; when this ''simplifying assumption'' is violated, e.g. in highly nonstationary or multimodal data, our conditional quantiles may become biased.

% Second, the number of bivariate dependence metrics evaluations can scale unfavorably by $\mathcal{O}(d^2)$, which can become a bottleneck for extremely high dimensions.

% Finally, our empirical study is limited to two low-to-moderate-dimensional regression benchmarks, leaving open questions about robustness to different data modalities (e.g. images or text), and deployment in streaming or online settings.

% Upcoming releases of \texttt{torchvinecopulib} would replace nested \texttt{Python} hashmaps with flat tensor views and add the vectorized union-find.
% Fully end-to-end differentiable learning of the VCG structure itself via gradient descent presents additional challenges and is an avenue for future research.

\newpage
\begin{ack}
	We thank Johannes Wiesel and Xianjin Yang for valuable discussions regarding bivariate dependence metrics, Xiaosheng You for helpful comments on the manuscript draft, and Saikiran Reddy Poreddy for assistance with Python scripting.
	We are also grateful to the anonymous reviewers for their constructive feedback that helped improve this paper.
	% This work was supported [or partially supported] by [Funding Agency Name(s), Grant Number(s), e.g., the National Research Foundation, Singapore under its AI Singapore Programme (AISG-GC-20XX-XXX)].
\end{ack}

% \section*{References}
% \bibliographystyle{apalike}
% \bibliographystyle{plainnat}
\bibliographystyle{plain}
\bibliography{references}% common bib file

\newpage
\appendix

% \section{Technical Appendices and Supplementary Material}
% Technical appendices with additional results, figures, graphs and proofs may be submitted with the paper submission before the full submission deadline (see above), or as a separate PDF in the ZIP file below before the supplementary material deadline. There is no page limit for the technical appendices.
\section{Pair copula construction and sampling in three dimensions} \label{sec:pcc}
% All headings should be lower case (except for first word and proper nouns), flush left, and bold.
% {\color{blue}Comment\\
% \begin{enumerate}
%     \item Sklar: mv distribution function $\to$ mv cop;
%     \item PCC/vine: mv cop $\to$ bicop, density (fit) and sim (inv Rosenblatt) in formula (low dimension);
%     \item vine construction in graph (without tree part)
%     \item full sampling in graph (given fitted vine)
%     \item cond sampling in graph (given fitted vine and source vertices)
% \end{enumerate}
% }
% 
By Sklar's theorem, for a $d$-dimensional random vector $\mathbf{X}\in\mathbb{R}^D$ with an absolutely continuous joint distribution function $F$, marginal distribution functions $F_d,~d\in\{0,\dots,d-1\}$, and an associated joint density function $f$, we have:

\begin{align*}
	F(x_0, \dots, x_{d-1}) & = C(F_0(x_0), \dots, F_{d-1}(x_{d-1}))
	\\
	f(x_0, \dots, x_{d-1}) & = c(F_0(x_0), \dots, F_{d-1}(x_{d-1}))
	\cdot
	f_0(x_0)
	\cdots
	f_{d-1}(x_{d-1})
\end{align*}

\noindent
where the $d$-dimensional copula $C$ is a multivariate distribution function on the $d$-dimensional hypercube $[0,1]^D$ with uniformly distributed marginals.
The corresponding copula density $c$ can be obtained by partial differentiation $c(u_0,\cdots,u_{d-1})=\frac{\partial^D}{\partial u_0\cdots\partial u_{d-1}}C(u_0,\cdots,u_{d-1})$ for all $\mathbf{u}=F(\mathbf{x})\in[0,1]^D$.
For absolutely continuous distributions, the copula $C$ is unique \citep{sklar_fonctions_1959}.
% The goal is to construct multivariate distributions using only bivariate building
% blocks. The appropriate tool to obtain such a construction is to use conditioning.
% Joe (1996) gave the first pair copula construction of a multivariate copula in terms
% of distribution functions, while Bedford and Cooke (2001, 2002) independently
% developed constructions expressed in terms of densities. Additionally they provided
% a general framework to identify all possible constructions.
% 

Focusing on dependence structures, consider the pair copula construction for a three-dimensional copula on continuous variables without marginals:

\begin{align*}
	c(u_0,u_1,u_2) & =
	c_{0,2|1}(C_{0|1}(u_0|u_1), C_{2|1}(u_2|u_1))
	~
	c_{1,2}(u_1,u_2)
	~
	c_{0,1}(u_0,u_1)
\end{align*}

\noindent
where
$c_{0,2|1}$ is the copula density associated with
the conditional bivariate copula $C_{0,2|1}(u_0,u_2|u_1)=C_{0,2|1}\left(
	C_{0|1}^{-1}(u_0|u_1),
	C_{2|1}^{-1}(u_2|u_1)
	~\big|u_1
	\right)$, which has a conditioned set $\{0,2\}$ and a conditioning set $\{1\}$ \citep{czado_analyzing_2019}.
Here $C_{0|1}(u_0|u_1)$ and $C_{2|1}(u_2|u_1)$ are conditional distribution functions associated with bivariate copulas $C_{0,1}$ and $C_{1,2}$, also known as h-functions and denoted as $h_{0|1,\emptyset}(u_0|u_1)=\frac{\partial C_{0,1}(u_0,u_1)}{\partial u_1}$ and $h_{2|1,\emptyset}(u_2|u_1)=\frac{\partial C_{1,2}(u_1,u_2)}{\partial u_1}$ \citep{aas_pair-copula_2009}.
The bivariate copula $C_{0,1}$ has a conditioned set $\{0,1\}$ and a conditioning set $\emptyset$, while the bivariate copula $C_{1,2}$ has a conditioned set $\{1,2\}$ and a conditioning set $\emptyset$.
Here we adopt the simplifying assumption that $c_{0,2|1}(u_0,u_2|x_1)=c_{0,2|1}(u_0,u_2)$, indicating that the bivariate copula associated with conditional distributions does not depend on specific values of the conditioning variables \citep{wattanawongwan_modelling_2023,nagler2025simplifiedvinecopulamodels}.

With this pair copula construction of $c$, we may sample from the trivariate copula using the inverse Rosenblatt transform. First, we sample $w_0,w_1,w_2\overset{i.i.d.}{\sim}Unif[0,1],~d=0,\dots,D$, then sequentially:

\begin{align*}
	u_0: & =w_0
	\\
	u_1: &
	=h^{-1}_{1|0,\emptyset}(w_1|u_0)
	\\
	u_2: &
	=h^{-1}_{2|1,\emptyset}\left(
	h^{-1}_{2|0,1}\left(w_2| h_{0|1,\emptyset}(u_0|u_1)\right)
	\big|u_1\right)
\end{align*}

\noindent
where $u_0, u_1, u_2$ is a sample from the tri-variate copula $c$, expressed in terms of $w_0, w_1, w_2$, with the sampling order (\cref{def:sample_order}, from right to left) denoted as $\mathcal{S}_\text{order}=(2,1,0)$.

A tri-variate copula has $3$ different pair copula constructions, each allowing for $4$ possible sampling orders.
An alternative inverse Rosenblatt transform for the sampling order $\mathcal{S}_\text{order}=(0,2,1)$ is:

\begin{align*}
	u_1: & =w_1
	\\
	u_2: &
	=h^{-1}_{2|1,\emptyset}(w_2|u_1)
	\\
	u_0: &
	=h^{-1}_{0|1,\emptyset}\left(
	h^{-1}_{0|2,1}\left(w_0| w_2\right)
	\big|u_1\right)
\end{align*}

\noindent
Notice both sampling orders require $3$ h-function-inverse calls, but the former has one additional h-function call, making the latter more efficient.
This pair copula construction and sampling method can scale to higher dimensions, though the formulas become increasingly complex.

\begin{definition}[Sampling order]
	\label{def:sample_order}
	An \emph{ordered} tuple $\mathcal{S}_\text{order}$, where $~0<|\mathcal{S}_\text{order}|\leq D$, is the sampling order of a $d$-dimensional vine on set $\mathcal{S}=\{0,1,\dots,d-1\}$. Its encoded set of variable vertices $\left\{
		\left\{\mathcal{S}_\text{order}[i]~|~\mathcal{S}_\text{order}[(i+1):|\mathcal{S}_\text{order}|]\cup\mathcal{S}\setminus\mathcal{S}_\text{order}\right\}
		~\bigg|~i\in\{0,\dots,|\mathcal{S}_\text{order}|-1\}
		\right\}
		\cup\left\{\{j|\emptyset\}~\bigg|~j\in\mathcal{S}\setminus\mathcal{S}_\text{order}\right\}$
	is a set of source vertices that allows sampling from the vine by \cref{algo:sim}.
	Here $\mathcal{S}_\text{order}[i]$ is the $i$-th element in $\mathcal{S}_\text{order}$, and $\mathcal{S}_\text{order}[(i+1):|\mathcal{S}_\text{order}|]$ represents the remaining successor elements in $\mathcal{S}_\text{order}$, with the $(i+1)$-th included but the $|\mathcal{S}_\text{order}|$-th excluded.
\end{definition}

\section{Vine copula and vine computational graph definitions} \label{sec:vine:cop:def}
% All headings should be lower case (except for first word and proper nouns), flush left, and bold.
To place the computational‑graph view in context, we first recall the classical notions of a vine (\cref{def:vine_mst}) and a vine copula (\cref{def:vinecop_mst}).
The decomposition of a $d$-dimensional ($d\geq3$) copula into a cascade of bivariate copulas is non-unique, but every possible decomposition can be represented by nested trees known as regular vine (R-vine) \citep{bedford_vines_new_2002}.
Each vine structure corresponds to a different decomposition of the joint copula density, with each tree representing a connected, undirected, acyclic graph $\bigl\{(\mathcal{V}_k, \mathcal{E}_k)\bigr\}_{k=0}^{d-2}$. Each vertex is a set of integers and each undirected edge connects two vertices.
\\
\begin{definition}[Vine]
	\label{def:vine_mst}
	A collection of trees $\bigl\{(\mathcal{V}_k, \mathcal{E}_k)\bigr\}_{k=0}^{d-2}$ is a $d$-dimensional vine on the set $\mathcal{V}_0=\{\{0\},\{1\},\dots,\{d-1\}\}$ if:

	\begin{enumerate}
		\item $(\mathcal{V}_0,\mathcal{E}_0)$ is a spanning tree,
		\item For $k\in\{1,\dots,d-2\}$, $(\mathcal{V}_k,\mathcal{E}_k)$ is a spanning tree with vertices $\mathcal{V}_k=\{a\cup b \big| \{a,b\}\in\mathcal{E}_{k-1}, |a\cup b|=k+1\}$, $|\mathcal{V}_k|=d-k$, $|\mathcal{E}_k|=d-k-1$,
		\item (Proximity condition)
		      Each undirected edge $e=\{a,b\}\in\mathcal{E}_k$ connects vertices $a,b\in\mathcal{V}_k$ where $|a\cap b|=k, |a\oplus b|=|(a\cup b)\setminus(a\cap b)|=2$, with $\oplus$ denoting the symmetric difference.
	\end{enumerate}
\end{definition}
\begin{definition}[Vine copula]
	\label{def:vinecop_mst}
	A vine copula $\bigl\{(\mathcal{V}_k,\mathcal{E}_k,\mathcal{C}_k)\bigr\}_{k=0}^{d-2}$ attach a bivariate copula to each undirected edge of a vine. For each undirected edge $e=\{a,b\}\in\mathcal{E}_{k}$:

	\begin{enumerate}
		\item The conditioning set of $e$ is $a\cap b$ with $|a\cap b|=k$,
		\item The conditioned set of $e$ is $a\oplus b$ with $|a\oplus b|=2$. The elements of this bivariate set $a\oplus b$ are denoted $l_i, r_i$ where $l_i<r_i$,
		\item The bivariate copula attached to $e$ is $C_{l_i, r_i| a\cap b}\in\mathcal{C}_k$,
		\item The corresponding conditional distribution functions (h-functions) are
		      $h_{l_i|r_i, a\cap b}(u_0|u_1)
			      =\frac{\partial}{\partial u_1}C_{l_i, r_i| a\cap b}(u_0,u_1)
		      $
		      and
		      $h_{r_i|l_i, a\cap b}(u_1|u_0)=
			      \frac{\partial}{\partial u_0}C_{l_i, r_i| a\cap b}(u_0,u_1)
		      $.
		      % \item the pseudo observations transformed by applying the h-functions of the bivariate copula attached to $e$ are 
		      % $u_{l_i|a\cap b \cup \{r_i\}}=h_{l_i|r_i,a\cap b}(u_{l_i|a\cap b}|u_{r_i|a\cap b})$ 
		      % and 
		      % $u_{r_i|a\cap b \cup \{l_i\}}=h_{r_i|l_i,a\cap b}(u_{r_i|a\cap b}|u_{l_i|a\cap b})$,
		      % % as separated by a vertical bar,
		      % where pseudo-observations $u_{\cdot|\cdots}$ could be recursively backtracked to observations $u_{d|\emptyset}=F_d(x_d), \{d\}\in\mathcal{V}_0$.
	\end{enumerate}
\end{definition}
\noindent
These objects describe which pairs of variables are coupled and which bivariate copulas are attached, but they remain purely combinatorial.
The next step—-and one of the key innovation of this paper—-is to make every data‑flow dependency explicit.
We introduce a DAG representation of vine (\cref{def:vine_dag}) and, with fitted bivariate models, into a vine copula  (\cref{def:vinecop_dag}).
\\
\begin{definition}[Vine Computational Graph (VCG)]
	\label{def:vine_dag}
	A VCG for a $d$-dimensional vine on the set $\{0,1,\dots,d-1\}$ consists of variable vertices, bivariate copula vertices, and directed edges connecting them, denoted as
	$\bigl\{(\mathcal{V}_k, \mathcal{E}_k, E_{k}^u, E_{k}^d)\bigr\}_{k=0}^{d-2}$ and satisfies the following conditions:
	\begin{enumerate}
		\item
		      $\mathcal{V}_0=\left\{
			      \{d|\emptyset\}|d\in\{0,1,\dots,d-1\}
			      \right\}$, where $\emptyset$ is the conditioning set,
		      % TODO: two bipartite, one tree, improve def
		\item
		      For $k\in\{0,1,\dots,d-2\}$, $(\mathcal{V}_k\cup\mathcal{E}_k\cup\mathcal{V}_{k+1}, E_{k}^u\cup E_{k}^d)$ forms a DAG,
		      where \raggedright\\
		      $\mathcal{E}_k\subseteq\left\{
			      \{l_i,r_i| \mathcal{S}_i\}
			      \big|~
			      \{
			      \{l_i|\mathcal{S}_i\},
			      \{r_i|\mathcal{S}_i\}
			      \}
			      \subseteq \mathcal{V}_{k},
			      \{l_i,r_i\}\subseteq
			      \mathcal{S}\setminus\mathcal{S}_i,
			      l_i<r_i
			      \right\}$,
		      $\mathcal{V}_{k+1}=\underset{e=\{l_i,r_i| \mathcal{S}_i\}\in\mathcal{E}_k}{\bigcup}
			      \left\{
			      \{l_i| \mathcal{S}_i\cup\{r_i\}\},
			      \{r_i|\mathcal{S}_i\cup\{l_i\}\}
			      % \big|~ e=\{l_i,r_i| \mathcal{S}_i\}
			      \right\}$,
		      $E_{k}^u=\underset{e=\{l_i,r_i| \mathcal{S}_i\}\in\mathcal{E}_k}{\bigcup}
			      \left\{
			      (\{l_i|\mathcal{S}_i\},e),
			      (\{r_i|\mathcal{S}_i\},e)
			      % \big|~e=\{l_i,r_i| \mathcal{S}_i\}
			      \right\}$,
		      $E_{k}^d=\underset{e=\{l_i,r_i| \mathcal{S}_i\}\in\mathcal{E}_k}{\bigcup}
			      \left\{
			      (e,\{l_i| \mathcal{S}_i\cup\{r_i\}\}),
			      (e,\{r_i|\mathcal{S}_i\cup\{l_i\}\})
			      \right\}$,
		      \\
		      $\mathcal{S}_i\subset\mathcal{S}$ is the conditioning set shared by both a bivariate copula vertex $e=\{l_i,r_i| \mathcal{S}_i\}$ as well as its two parent variable vertices $\{l_i|\mathcal{S}_i\}$ and $\{r_i|\mathcal{S}_i\}$,

		\item
		      The skeleton (the undirected graph formed by removing direction of all edges in the DAG) of $(\mathcal{V}_{0}\cup\mathcal{E}_{0}, ~E_{0}^u)$ forms a spanning tree,
		      \label{def:vine_dag:tree_0}

		\item
		      For $k\in\{1,2,\dots,d-2\}$, the skeleton of $(\mathcal{E}_{k-1}\cup\mathcal{V}_{k}\cup\mathcal{E}_{k},~E_{k-1}^d\cup E_{k}^u)$ forms a spanning tree.
		      \label{def:vine_dag:tree_k}

		      % A vine copula $\{(\mathcal{V}_k,\mathcal{E}_k,\mathcal{C}_k)|k\in\{0,1,\dots,d-2\}\}$ attaches a bivariate copula to each undirected edge of a vine, such that for each undirected edge $e=\{a,b\}\in\mathcal{E}_{k}$ connecting vertices $a$ with $b$:
	\end{enumerate}
\end{definition}

\begin{definition}[Vine copula]
	\label{def:vinecop_dag}
	A vine copula
	$\bigl\{ (\mathcal{V}_k, \mathcal{E}_k, \mathcal{C}_{k}, E_{k}^u, E_{k}^d)\bigr\}_{k=0}^{d-2}$ attaches a bivariate copula model $C_{l_i,r_i| \mathcal{S}_i}(\cdot,\cdot)\in\mathcal{C}_k$ to each bivariate copula vertex $e=\{l_i,r_i| \mathcal{S}_i\}\in\mathcal{E}_k$ in the computational graph of a vine,
	\begin{enumerate}
		\item the h-functions (also known as conditional distribution functions) of the bivariate copula attached to $e$ are
		      $h_{l_i|r_i, \mathcal{S}_i}(u_0|u_1)=
			      \frac{\partial}{\partial u_1}C_{l_i,r_i| \mathcal{S}_i}(u_0,u_1)=
			      C_{l_i| \mathcal{S}_i\cup\{r_i\}}(u_0|u_1)
		      $
		      and
		      $h_{r_i|l_i, \mathcal{S}_i}(u_1|u_0)=
			      \frac{\partial}{\partial u_0}C_{l_i,r_i| \mathcal{S}_i}(u_0,u_1)=
			      C_{r_i|\mathcal{S}_i \cup \{l_i\}}(u_1|u_0)
		      $,

		\item the pseudo-observations generated by applying the h-functions of the bivariate copula attached to $e$ are
		      $u_{l_i| \mathcal{S}_i\cup\{r_i\}}=h_{l_i|r_i,\mathcal{S}_i}(u_{l_i|\mathcal{S}_i}|u_{r_i|\mathcal{S}_i})$
		      and
		      $u_{r_i|\mathcal{S}_i \cup \{l_i\}}=h_{r_i|l_i,\mathcal{S}_i}(u_{r_i|\mathcal{S}_i}|u_{l_i|\mathcal{S}_i})$,
		      by which pseudo-observations $u_{\cdot|\cdot}$ can be recursively computed based on observations $u_{v}, v\in\mathcal{V}_0$.
	\end{enumerate}
\end{definition}

\section{
  Constructing C-vines and D-vines admitting customized conditioning variables
 } \label{sec:dcvine}
% All headings should be lower case (except for first word and proper nouns), flush left, and bold.
% 
\begin{figure}
	\centering
	\includegraphics[width=0.8\textwidth]{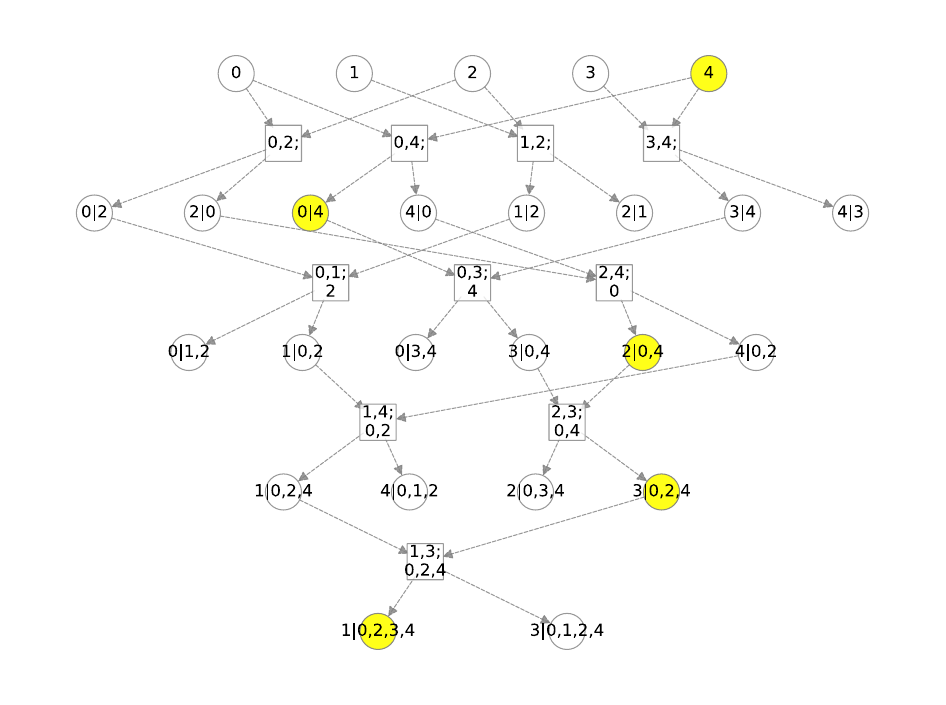}
	\caption{
		D-vine computational graph ($d{=}5$) with sampling order $(1,3,2,0,4)$.
	}
	\label{fig:dvine_dag}
\end{figure}
\begin{figure}
	\centering
	\includegraphics[width=0.8\textwidth]{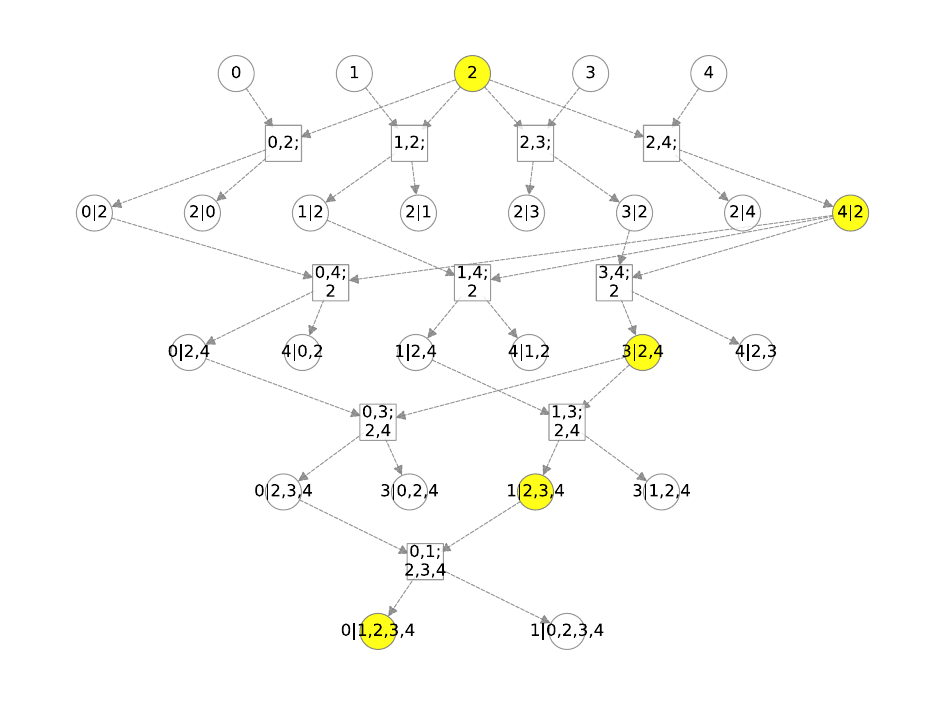}
	\caption{
		C-vine computational graph ($d{=}5$) with sampling order $(0,1,3,4,2)$.
	}
	\label{fig:cvine_dag}
\end{figure}
% 
% \subsection{C-vines and D-vines conditional sampling}
% 
% D-vine sampling almost always calls h-functions, while C-vine sampling can avoid calling h-function.
% 
% In \cref{fig:dvine_dag}, first we visit the vertex $\{4\}$ which is both a source and a target variable vertex. 
% Secondly, we begin with the source $\{0|4\}$, which calls the h-function-inverse of bivariate copula $\{0,4\}$ onto $\{0|4\}$ and $\{4\}$ to walk the path $\{0|4\}\to\{4\}$.
% Thirdly we begin with the source $\{2|{0,4}\}$. The step $\{2|{0,4}\}\to\{2|0\}$ calls a h-function-inverse of the bivariate copula $\{2,4|0\}$ that needs the other parent variable $\{4|0\}$, which is visited by applying a h-function of bivariate copula $\{0,4\}$ onto its parents $\{0\}$ and $\{4\}$. The next step $\{2|0\} \to \{2\}$ calls a h-function-inverse of the bivariate copula $\{0,2\}$ that requires its parent $\{0\}$, which is already visited in the previous path $\{0|4\} \to \{0\}$.
% 
% Take the deepest source vertex in \cref{fig:cvine_dag} as an example, we walk the path: $\{0|1,2,3,4\} \to \{0|2,3,4\} \to \{0|2,4\} \to \{0|2\} \to \{0\}$, during which its conditioning set is gradually depleted. Along this path, all upper vertices including $\{1|2,3,4\},~\{3|2,4\},~\{4|2\},~\{2\}$ are highlighted as source vertices and are already visited before walking this path and thus no h-function is called.
% 
% \noindent\\
To accommodate the structure constraints of C-vines and D-vines, we adapt the bivariate copula vertex selection step in \cref{algo:vine} from \cref{algo:vine:kruskal:init:first} to \cref{algo:vine:kruskal:last}.
Constructing a D-vine with a given set of conditioning variables $\mathcal{S}_\text{cond}$, amounts at solving a traveling salesman problem \cite[TSP, see e.g. ][]{lawler1985traveling,gutin2006tsp} at level $k=0$ with a precedence constraint. In this TSP, the Hamiltonian cycle on $\mathcal{V}_0=\{\{l|\emptyset\}|l\in \{0, \dots, d-1 \}\}$ ensures that the vertices corresponding to $\mathcal{S}_\text{cond}$ are clustered together.
For each undirected edge $e=\{\{l_i|\emptyset\},\{r_i|\emptyset\}\}$, its travel cost is represented as $\log(1+\frac{1}{w[e]+\epsilon})$, where $w[e]$ denotes the absolute value of the bivariate dependence metric, and $\epsilon>0$ is to maintain the denominator positive.
To obtain the Hamiltonian path, we drop one undirected edge outside the cluster of $\mathcal{S}_\text{cond}$ to minimize the total travel cost.
Due to proximity conditions, the entire structure of the D-vine is determined at the $0$-th level.
The number of bivariate dependence metric calculations is of $\mathcal{O}(d^2)$ at the $0$-th level, and reduces to $\mathcal{O}(d)$ at subsequent levels (which could be omitted since the structure is known then).
To construct a C-vine with a given set of conditioning variables $\mathcal{S}_\text{cond}$,
we select the central variable vertex at each level based on the maximum sum of absolute bivariate dependence metrics to all other variable vertices, as stored in $w$.
At levels $k\in\{0,1,\dots,|\mathcal{S}_\text{cond}|-1\}$, the central variable vertex is restricted to those whose conditioned set is a subset of $\mathcal{S}_\text{cond}$.
At levels $k\in\{|\mathcal{S}_\text{cond}|,\dots,d-2\}$, the central variable vertex is restricted to those whose conditioned set is a subset of $\{0, \dots, d-1 \}\setminus\mathcal{S}_\text{cond}$.
The central vertices may need to be determined repeatedly at each level.
Since at least half of the variable vertices at each level share the same conditioning set, the number of bivariate dependence metric calculations remains $\mathcal{O}(d^2)$ at all levels.

\section{Algorithms} \label{sec:algo}
\cref{algo:sim} can be described as follow.
We first seed the graph with pseudo-observations at source vertices.
Sequentially from the shallowest to the deepest level, we start from the $d$ source vertices to walk along $d$ unique upward paths until reaching their corresponding $d$ target vertices at the top level, which keep the same singleton conditioned sets but empty conditioning sets, as detailed from \cref{algo:sim:main:first} to \cref{algo:sim:main:last}.
The $d$ source vertices are arranged such that their paths by h-function-inverse do not intersect at any bivariate copula vertex.
Along each path, iteratively we use the h-function-inverse of a bivariate copula for the inverse Rosenblatt transform to ascend from a child variable vertex to visit its parent variable vertex with the same conditioned set via an upward visit.
During this process, we may (recursively) request h-functions for the other upper (``opposite" parent) variable vertex as a downward visit, if it is not yet visited, as detailed from \cref{algo:sim:visit:first} to \cref{algo:sim:visit:last}. 
Conditional sampling, as described in \cref{sec:sampling}, is obtained by modifying \cref{algo:sim:main:cond:first} to \cref{algo:sim:main:cond:last} to record the $|\mathcal{S}_\text{cond}|$ variable vertices with empty conditioning set at the top level as visited source vertices.
\newpage
\begin{algorithm}
	\caption{Sample from vine copula by inverse Rosenblatt transform} \label{algo:sim}
	\begin{algorithmic}[1]
		\footnotesize
		\Require
		$\bigl\{(\mathcal{V}_k, \mathcal{E}_k, \mathcal{C}_k, E_{k}^u, E_{k}^d)\bigr\}_{k=0}^{d-2}$ vine copula,
		$\mathcal{V}_\text{source} \subset \bigcup_{k}{\mathcal{V}_k}
			% \left\{
			% \{l|\mathcal{S}_i\} \big| v\in\mathcal{S}\setminus\mathcal{S}_{cond}, 
			% \mathcal{S}_{cond}\subset\mathcal{S}
			% \right\}
		$ source vertices, and $|\mathcal{V}_\text{source}|=d$
		\Ensure $u_0,\dots,u_{d-1}$ compose a sample from the vine copula
		% with bivariate copula in $\mathcal{F}$ at each level 
		% for conditional sampling from $\mathcal{S}_{cond}$
		\Statex
		\Statex \textbf{Helper Functions for Graph Traversal}
		\Function{Visit}{$l$, $\mathcal{S}$, $isupward$}
		\label{algo:sim:visit:first}
		% \State{$\mathcal{S}_{cond}^{up} \gets \mathcal{S}_i \cup \{v\}$}
		\For{$\{l_i,r_i| \mathcal{S}_i\} \gets \mathcal{E}_{\left|\mathcal{S}\right|-1}$}
		\If{$\{l_i,r_i\} \cup \mathcal{S}_i = \{l\} \cup \mathcal{S}$}
		\State{\textbf{break}}
		\Comment{locate parent bivariate copula vertex}
		\EndIf
		\EndFor
		% \Comment{depth-first search on binary tree}
		\State{$l_{oppo}\gets l_i$}
		\Comment{locate the opposite parent variable}
		\If{$l=l_i$}
		\State{$l_{oppo}\gets r_i$}
		\EndIf
		\State{$tovisit\gets \{
				\{l_i| \mathcal{S}_i\},
				\{r_i| \mathcal{S}_i\}
				\}$}
		\Comment{check parent variable availability}
		\If{$isupward$}
		\State{$tovisit\gets \{\{l_{oppo}| \mathcal{S}_i\}\}$}
		\EndIf
		\For{$\{l_i| \mathcal{S}_i\} \gets tovisit$}
		\If{$\{l_i| \mathcal{S}_i\} \notin V$}
		\State{\Call{Visit}{$l_i,\mathcal{S}_{i}, False$}}
		\EndIf
		\EndFor
		\If{$isupward$}
		\Comment{h-function-inverse, upward}
		\State{
		$u_{l|\mathcal{S}_i}\gets
			h^{-1}_{l|l_{oppo},\mathcal{S}_i}
			\left(u_{l|\mathcal{S}_i\cup\{l_{oppo}\}} \bigg|u_{l_{oppo}|\mathcal{S}_i}
			\right)$}
		% \EndIf
		\State{$V\gets V\cup \{ \{l|\mathcal{S}_i\} \}$}
		\State{\Return $\{l|\mathcal{S}_i\}$}
		\Else
		\Comment{h-function, downward}
		\State{
		$u_{l|\mathcal{S}_i\cup\{l_{oppo}\}}\gets
			h_{l|l_{oppo},\mathcal{S}_i}
			\left(u_{l|\mathcal{S}_i} \bigg|u_{l_{oppo}|\mathcal{S}_i}
			\right)$
		}
		\State{$V\gets V\cup \{ \{l|\mathcal{S}_i\cup\{l_{oppo}\} \} \}$}
		\EndIf
		\EndFunction
		\label{algo:sim:visit:last}
		\Statex
		\Statex \textbf{Sequential Paths from Sources to Targets}
		\State{$V\gets\emptyset$}
		\Comment{record for visited variable vertices}
		\label{algo:sim:main:first}
		\label{algo:sim:main:cond:first}
		\For{$\{l|\mathcal{S}\}\gets\mathcal{V}_\text{source}$ sorted in ascending order of $|\mathcal{S}|$}
		\Comment{from shallowest to deepest}
		\State{sample $u_{l|\mathcal{S}}\sim U[0,1]$}
		\State{$V\gets V \cup \{\{l|\mathcal{S}\}\}$}
		\label{algo:sim:main:cond:last}
		\State{$\{l_{next}|\mathcal{S}_{l_{next}}\}\gets \{l|\mathcal{S}\}$}
		% \State{$\mathcal{S}_{v_{next}}\gets $}
		\While{$|\mathcal{S}_{l_{next}}|>0$}
		\State{
		$\{l_{next}|\mathcal{S}_{l_{next}}\}\gets$
		\Call{Visit}{$l_{next}, \mathcal{S}_{l_{next}}, True$}
		}
		\EndWhile
		\State{$u_l\gets u_{l|\emptyset}$}
		\EndFor
		\label{algo:sim:main:last}
	\end{algorithmic}
\end{algorithm}
\newpage
\begin{algorithm}
	\caption{Schedule sampling order and source vertices} \label{algo:source}
	\begin{algorithmic}[1]
		\footnotesize
		\Require $\mathcal{S}_\text{cond}\subset\{0,1,\dots,d-1\}
		$ set of conditioning variables,
		$\bigl\{(\mathcal{V}_k, \mathcal{E}_k, \mathcal{C}_k, E_{k}^u, E_{k}^d)\bigr\}_{k=0}^{d-2}$ vine copula
		% $\{ (\mathcal{V}_k, \mathcal{E}_k, E_{k}^u, E_{k}^d)| k\in\{0,1,\dots,d-2\}\}$
		% , $V$ visit records, $\mathcal{V}$ temporary source vertices
		\Ensure $\mathcal{S}_\text{order}$ sampling order, $\mathcal{V}_\text{source}$ source vertices
		\Statex
		\Statex \textbf{Helper Functions for Counting}
		\Function{GetSource}{$\mathcal{S}_\text{order}'$}
		\Comment{infer source vertices given sampling order}
		\State{$\mathcal{V}\gets\emptyset$}
		\Comment{list to store source vertices}
		\For{$i\in \{0,\dots,|\mathcal{S}_\text{order}'|-1\}$}
		\Comment{implied source vertices, bottom-up}
		\State{$\mathcal{V}\gets \mathcal{V}\cup\left\{\{
			\mathcal{S}_\text{order}'[i]|
			\mathcal{S}_\text{cond} \cup
			\{\mathcal{S}_\text{order}'[(i+1):|\mathcal{S}_\text{order}'|]\}
			\}\right\}$}
		\EndFor
		\For{$i\in \mathcal{S}\setminus\mathcal{S}_\text{order}'$}
		\Comment{other source vertices, on top}
		\State{$\mathcal{V}\gets \mathcal{V}\cup\left\{\{i|\emptyset\}\right\}$}
		\EndFor
		\State{\Return $\mathcal{V}$}
		\EndFunction
		\Statex
		\Function{Query}{$\mathcal{S}_\text{order}'$}
		\Comment{count number of h-function calls given sampling order}
		\label{algo:source:count:first}
		% \Function{Visit}{$v,\mathcal{S}_{v}, isupward$}
		% \label{algo:source:visit:first}
		% \For{$\{l_i,r_i| \mathcal{S}_i\} \gets \mathcal{E}_{\left|\mathcal{S}_{v}\right|-1}$}
		% \If{$\{l_i,r_i\} \cup \mathcal{S}_i = \{l\} \cup \mathcal{S}$}
		%     \State{\textbf{break}}
		%     \Comment{locate parent bivariate copula vertex}
		% \EndIf
		% \EndFor
		% \State{$v_{oppo}\gets l_i$}
		% \Comment{locate the opposite parent variable}
		% \If{$v=l_i$}
		% \State{$v_{oppo}\gets r_i$}
		% \EndIf
		% \State{$tovisit\gets \{
		% \{l_i| \mathcal{S}_i\}, 
		% \{r_i| \mathcal{S}_i\}
		% \}$}
		% \Comment{check parent variable availability}
		% \If{$isupward$}
		% \State{$tovisit\gets \{\{v_{oppo}| \mathcal{S}_i\}\}$}
		% \EndIf
		% \For{$\{|_i| \mathcal{S}_i\} \gets tovisit$}
		% \If{$\{|_i| \mathcal{S}_i\} \notin V$}
		% \State{\Call{Visit}{$v',\mathcal{S}_{v'}, False$}}
		% \State{$n_h\gets n_h+1$}
		% \Comment{count}
		% \EndIf
		% \EndFor
		% \State{$tovisit\gets \{
		% \{l_i| \mathcal{S}_i\}, 
		% \{r_i| \mathcal{S}_i\},
		% \{l| \mathcal{S}_{v}\}
		% \}$}
		% \For{$\{|_i| \mathcal{S}_i\} \gets tovisit$}
		% \State{$V\gets V\cup \{|_i| \mathcal{S}_i\}$}
		% \EndFor
		% \If{$isupward$}
		% \State{\Return $\{l|\mathcal{S}_i\}$}
		% \EndIf
		% \EndFunction
		% \label{algo:source:visit:last}
		% 
		% \Statex
		% \State{$n_h\gets0,~ V\gets\emptyset,~ \mathcal{V}\gets\emptyset$}
		% 
		\State{
			$\mathcal{V}\gets$
			\Call{GetSource}{$\mathcal{S}_\text{order}'$}}
		\State{
		$n_h\gets$
		number of times \textsc{Visit}$(\ast,\ast,\mathrm{False})$ is executed when Algorithm~\ref{algo:sim} runs on the vine copula
		$\bigl\{(\mathcal{V}_k, \mathcal{E}_k, \mathcal{C}_k, E_{k}^u, E_{k}^d)\bigr\}_{k=0}^{d-2}$
		with source vertices \(\mathcal{V}\) and number of dimension $d$.
		% n_h\gets
		% number of times \textsc{Visit}{,,$(\cdot,\cdot,isupward=\mathrm{False})$} invocations, denoted as $n_h$, using $\mathcal{S},~D,~\mathcal{V},~\{(\mathcal{V}_k, \mathcal{E}_k, \mathcal{C}_k, E_{k}^u, E_{k}^d)| k\in\{0,1,\dots,d-2\}\}$ for \cref{algo:sim}
		}
		% \For{$\{l|\mathcal{S}_{v}\}\gets\mathcal{V}$ sorted in ascending order of $|\mathcal{S}_{v}|$}
		% \Comment{from top to bottom}
		% \State{$V\gets V \cup \{\{l|\mathcal{S}_{v}\}\}$}
		% % \label{algo:sim:main:cond:last}
		% \State{$\{v_{next}|\mathcal{S}_{v_{next}}\}\gets \{l|\mathcal{S}_{v}\}$}
		% \While{$|\mathcal{S}_{v_{next}}|>0$}
		% \State{
		% $\{v_{next}|\mathcal{S}_{v_{next}}\}\gets$
		% \Call{Visit}{$v_{next}, \mathcal{S}_{v_{next}}, True$}
		% }
		% \EndWhile
		% \EndFor
		\State{\Return $n_h$}
		\EndFunction
		\label{algo:source:count:last}
		\Statex
		\Statex\textbf{Sequential Greedy Selection from Bottom to Top}
		\State{$\mathcal{S}_\text{order}\gets\emptyset$}
		\label{algo:source:select:first}
		\While{$d-|\mathcal{S}_\text{order}|>1$}
		\For{$\{l_i,r_i| \mathcal{S}_i\} \gets \mathcal{E}_{d-2-|\mathcal{S}_\text{order}|}$}
		\If{$l_i\notin\mathcal{S}_\text{order}~\mathbf{and}~r_i\notin\mathcal{S}_\text{order}$}
		\Comment{locate the unvisited bivariate copula vertex}
		\If{$\left|\{l_i,r_i\}\cap(\{0, \dots, d-1 \}\setminus\mathcal{S}_\text{cond})\right|=1$}
		\Comment{prioritize the one in $\{0, \dots, d-1 \}\setminus\mathcal{S}_\text{cond}$}
		\State{$\mathcal{S}_\text{order}\gets\mathcal{S}_\text{order}\cup(\{l_i,r_i\}\setminus\mathcal{S}_\text{cond})$}
		\ElsIf{
			\Call{Query}{$\mathcal{S}_\text{order}\cup\{l_i\}$}
			$\leq$
			\Call{Query}{$\mathcal{S}_\text{order}\cup\{r_i\}$}
		}
		\State{$\mathcal{S}_\text{order}\gets\mathcal{S}_\text{order}\cup\{l_i\}$}
		\Comment{select $l_i$ when in tie}
		\Else
		\State{$\mathcal{S}_\text{order}\gets\mathcal{S}_\text{order}\cup\{r_i\}$}
		\EndIf
		\EndIf
		\EndFor
		\If{$|\mathcal{S}_\text{order}|=|\{0, \dots, d-1 \}\setminus\mathcal{S}_\text{cond}|$}
		\State{\textbf{break}}
		\EndIf
		\EndWhile
		\If{$|\mathcal{S}_\text{order}|<|\{0, \dots, d-1 \}\setminus\mathcal{S}_\text{cond}|$}
		\State{$\mathcal{S}_\text{order}\gets\mathcal{S}_\text{order}\cup(\{0, \dots, d-1 \}\setminus\mathcal{S}_\text{cond}\setminus\mathcal{S}_\text{order})$}
		\EndIf
		% \State{$\mathcal{S}_{order}\gets\mathcal{S}_{order}\cup(\mathcal{S}\setminus\mathcal{S}_{order})$}
		\label{algo:source:select:last}
		\State{
			$\mathcal{V}_\text{source}\gets\Call{GetSource}{\mathcal{S}_\text{order}}$
		}
		% 
		% \For{$i\in \{0,\dots,|\mathcal{S}_{order}|-1\}$}
		% \label{algo:source:order2source:first}
		% \State{$\mathcal{V}_{source}\gets \mathcal{V}_{source}\cup\left\{\left\{
		% \mathcal{S}_{order}[i]|
		% \mathcal{S}_{cond}\cup
		% \left\{\mathcal{S}_{order}\left[(i+1):|\mathcal{S}_{order}|\right]\right\}
		% \right\}\right\}$}
		% \Comment{implied source vertices, bottom-up}
		% \EndFor
		% \For{$i\in \mathcal{S}_{cond}$}
		% \State{$\mathcal{V}_{source} \gets \mathcal{V}_{source} \cup \left\{\{i|\emptyset\}\right\}$}
		% \Comment{other source vertices, on top}
		% \EndFor
		\label{algo:source:order2source:last}
	\end{algorithmic}
\end{algorithm}
\newpage
\begin{algorithm}
	\caption{Construct an R-vine admitting customized conditioning variables} \label{algo:vine}
	\begin{algorithmic}[1]
		\footnotesize
		\Require $\mathcal{S}_\text{cond}\subset \{0,1,\dots,d-1\}, $
		% $\mathcal{P}$ roots array, $\mathcal{R}$ ranks array, 
		$f(\cdot,\cdot)$ bivariate dependence metric function, $w$ absolute values of bivariate dependence metric function
		\Ensure The R-vine $\{(\mathcal{V}_k, \mathcal{E}_k, \mathcal{C}_k)|k\in\{0,1,\dots,d-2\}\}$
		% with bivariate copula in $\mathcal{F}$ at each level 
		for conditional sampling from $\mathcal{S}_\text{cond}$
		% 
		% \Statex
		% \Statex \textbf{Helper Functions for Disjoint-Set}
		% \Function{Find}{$x$}
		% \label{algo:vine:find:first}
		%     \Comment{path compression}
		%     \If{$\mathcal{P}[x] \neq x$}
		%         \State{$\mathcal{P}[x] \gets \Call{Find}{\mathcal{P}[x]}$}
		%     \EndIf
		%     \State{\Return $\mathcal{P}[x]$}
		% \EndFunction
		% \label{algo:vine:find:last}
		% % 
		% \Function{Union}{$x, y$}
		% \label{algo:vine:union:first}
		%     \Comment{union by rank}
		%     \State{$rootX \gets \Call{Find}{x}$}
		%     \State{$rootY \gets \Call{Find}{y}$}
		%     \If{$\mathcal{R}[rootX] < \mathcal{R}[rootY]$}
		%         \State{$\mathcal{P}[rootX]\gets rootY$}
		%     \Else
		%         \State{$\mathcal{P}[rootY]\gets rootX$}
		%         \If{$\mathcal{R}[rootX] = \mathcal{R}[rootY]$}
		%             \State{$\mathcal{R}[rootX] \gets \mathcal{R}[rootX] + 1$}
		%         \EndIf
		%     \EndIf
		% \EndFunction
		% \label{algo:vine:union:last}
		% 

    \Statex
    \Statex \textbf{Helper Functions for Disjoint-Set}
    \Function{Find}{$x$}
    \label{algo:vine:find:first}
        \Comment{path compression}
        \If{$\mathcal{P}[x] \neq x$}
            \State{$\mathcal{P}[x] \gets \Call{Find}{\mathcal{P}[x]}$}
        \EndIf
        \State{\Return $\mathcal{P}[x]$}
    \EndFunction
    \label{algo:vine:find:last}
    \Function{Union}{$x, y$}
    \label{algo:vine:union:first}
        \Comment{union by rank}
        \State{$rootX \gets \Call{Find}{x}$}
        \State{$rootY \gets \Call{Find}{y}$}
        \If{$\mathcal{R}[rootX] < \mathcal{R}[rootY]$}
            \State{$\mathcal{P}[rootX]\gets rootY$}
        \Else
            \State{$\mathcal{P}[rootY]\gets rootX$}
            \If{$\mathcal{R}[rootX] = \mathcal{R}[rootY]$}
                \State{$\mathcal{R}[rootX] \gets \mathcal{R}[rootX] + 1$}
            \EndIf
        \EndIf
    \EndFunction
    \label{algo:vine:union:last}

		\Statex
		\For {$k \gets \{0,\dots,d-2\}$}
		\If{$k=0$}
		\label{algo:vine:0:first}
		% \For{$v\gets\{0,\dots,d-1\}$}
		\State{$\mathcal{V}_0 \gets \{\{l|\emptyset\} | v\in\mathcal{S}\}$}
		\Comment{top level variable vertices}
		% \EndFor
		\EndIf
		\label{algo:vine:0:last}
		\Statex \textbf{List Bivariate Copula Vertex Candidates}
		\State{$\mathcal{E}\gets\emptyset$, $w\gets\emptyset$}
		\label{algo:vine:edgelist:first}
		% \Comment{list of bivariate copula vertex candidates}
		\For{$
				\left(\{l_i|\mathcal{S}_{i}\},~
				\{r_j|\mathcal{S}_{j}\}\right)
				\gets
				\left\{
				\left(\{l_i|\mathcal{S}_{i}\},~
				\{r_j|\mathcal{S}_{j}\}\right)
				\big|
				\{\{l_i|\mathcal{S}_{i}\},
				\{r_j|\mathcal{S}_{j}\}\}
				\subseteq\mathcal{V}_k,
				i<j
				\right\}
			$
		}
		% proximity conditions
		\If{$\mathcal{S}_{i}=\mathcal{S}_{j}$}
		\Comment{proximity condition}
		\label{algo:vine:prox_cond_check}
		\State{$\mathcal{E} \gets \mathcal{E}\cup\{\{l_i,r_i| \mathcal{S}_{i}\}\}$}
		\State{$w[\{l_i,r_i| \mathcal{S}_{i}\}] \gets|f
				\left(\{l_i| \mathcal{S}_{i}\},~
				\{r_i| \mathcal{S}_{i}\}\right)|$}
		\Comment{bivariate dependence metric}
		\label{algo:vine:bidep}
		\EndIf
		\EndFor
		\label{algo:vine:edgelist:last}
		\Statex \textbf{Select Bivariate Copula Vertices}
		\State{$\mathcal{P}\gets\emptyset$, $\mathcal{R}\gets\emptyset$}
		\Comment{
			a \emph{disjoint-set} with roots and ranks for path-compression and union-by-rank
		}
		\label{algo:vine:kruskal:init:first}
		\State{$\mathcal{E}_k\gets\emptyset$}
		\Comment{edge list to store results}
		\For{$\{l|\mathcal{S}\}\gets\mathcal{V}_{k}$}
		\State{$\mathcal{P}[\{l|\mathcal{S}\}] \gets \mathcal{S} \cup \{l\}$}
		\label{algo:vine:root}
		\State{$\mathcal{P}[\{\mathcal{S} \cup \{l\}\}] \gets \mathcal{S} \cup \{l\}$}
		\State{$\mathcal{R}[\{l|\mathcal{S}\}] \gets 0
			$}
		% add $l|\mathcal{S}_{cond}^{d}$ to $\mathcal{S}_{disjoint}$,
		% mark its parent as $\{v\}\cup\mathcal{S}_{cond}^{d}$
		\EndFor
		\State{$\mathcal{E}_{s1}\gets\left\{\{l_i,r_i| \mathcal{S}_i\} \big|
				\{l_i,r_i\}\subseteq\mathcal{S}_\text{cond}, ~
				\{l_i,r_i|\mathcal{S}_{i}\} \in \mathcal{E}
				\right\}$}
		\State{$\mathcal{E}_{s2} \gets \mathcal{E}\setminus\mathcal{E}_{s1}$}
		\label{algo:vine:kruskal:init:last}
		\For{$\mathcal{E} \gets \{\mathcal{E}_{s1}, \mathcal{E}_{s2}\}$}
		\label{algo:vine:kruskal:first}
		\For{$\{l_i,r_i| \mathcal{S}_i\} \gets \mathcal{E}$ sorted by $w[\{l_i,r_i| \mathcal{S}_i\}]$ in descending order}
		% \Comment{priority queue}
		\If{
			\Call{Find}{$\{l_i|\mathcal{S}_i\}$}
			$\neq$
			\Call{Find}{$\{r_i|\mathcal{S}_i\}$}
		}
		\State{$\mathcal{E}_k \gets \mathcal{E}_k\cup\{\{l_i,r_i| \mathcal{S}_i\}\}$
		}
		\State{\Call{Union}{$\{l_i|\mathcal{S}_i\}$, $\{r_i|\mathcal{S}_i\}$}}
		\EndIf
		\EndFor
		\EndFor
		\label{algo:vine:kruskal:last}
		\Statex \textbf{Truncate or Fit Bivariate Copula Models and Rosenblatt Transform}
		\State{$\mathcal{C}_{k}\gets\emptyset$, $\mathcal{V}_{k+1}\gets\emptyset$}
		\label{algo:vine:est_sel_trunc:first}
		\For{$\{l_i,r_i| \mathcal{S}_i\}\gets \mathcal{E}_k$}
		\State{Use (pseudo-)observations at parent variable vertices $\{l_i| \mathcal{S}_i\}$ and $\{r_i| \mathcal{S}_i\}$ to fit the bivariate copula model for $C_{l_i,r_i| \mathcal{S}_i}$, and truncate to the Independence copula when necessary.}
		\State{Use the h-function of  $C_{l_i,r_i| \mathcal{S}_i}$ to compute pseudo-observations for the child variable vertices $\{l_i| \mathcal{S}_i\cup\{r_i\}\}$ and $\{r_i| \mathcal{S}_i\cup\{l_i\}\}$.}
		\State{$\mathcal{C}_k \gets \mathcal{C}_k \cup \left\{C_{l_i,r_i| \mathcal{S}_i}\right\}$}
		\State{$\mathcal{V}_{k+1} \gets \mathcal{V}_{k+1} \cup \left\{\{l_i| \mathcal{S}_i\cup\{r_i\}\}, \{r_i| \mathcal{S}_i\cup\{l_i\}\} \right\}$}
		\EndFor
		\label{algo:vine:est_sel_trunc:last}
		\EndFor
	\end{algorithmic}
\end{algorithm}

\newpage
\subsection{Optimality of \cref{algo:source} in greedoid language} \label{sec:sample_order:optimal}
% All headings should be lower case (except for first word and proper nouns), flush left, and bold.
% \TYcom{
% comment: search problem and algo, greeedoid language, proof
% \\}
We wish to show that the bottom-up greedy procedure in \cref{algo:source} finds a sampling order that minimizes the total number of h-function calls in \cref{algo:sim}. We cast the space of all valid sampling orders as a \emph{greedoid language}, then invoke the sufficient conditions under which a greedy algorithm on greedoid is guaranteed to produce a globally optimal base.

\subsubsection{The search problem}
A \emph{sampling order} is a permutation of the variables $\mathcal{S}=\{0,1,\dots,d-1\}$ compatible with the vine structure, and its cost is the number of h-function evaluations induced when we sample from the vine in that order. Building it up one variable at a time is naturally viewed as a search in the tree of all sampling orders.
In this tree, nodes represent sampling orders, and edges represent the addition of a variable to the order.

\subsubsection{Greedoid language preliminaries}
The following notations are used following conventions: with $S$ being a ground set, for a subset $X\subseteq S$ and an element $x\in S$ we write $X+x$ and $X-x$ instead of $X\cup \{x\}$ and $X\setminus\{x\}$, respectively.

Recall from the theory of greedoids \citep{korte1984greedoids, korte_greedoids_1991}:

\begin{definition}[Greedoid]\label{def:greedoid}
	A greedoid $G=(S,\mathcal{F})$ is a pair consisting of a finite ground set $S$ and a collection of its subsets $\mathcal{F}\subseteq 2^S$ such that the following properties are fulfilled:
	\begin{enumerate}
		\item $\emptyset \in \mathcal{F}$
		\item \label{def:greedoid:2} if $X,Y\in\mathcal{F}$ and $|X|<|Y|$ then there exists a $y\in Y- X$ such that $X+y\in\mathcal{F}$
	\end{enumerate}
\end{definition}
\noindent
Members of $\mathcal{F}$ are called feasible sets.

\emph{Minors of greedoids} are defined as follows:
If $G=(S,\mathcal{F})$ is a greedoid and $X\subseteq S$ is an arbitrary subset then the \emph{deletion} of $X$ yields the greedoid $G\setminus X=(S-X,\mathcal{F}\setminus X)$, where $\mathcal{F}\setminus X=\{Y\subseteq S-X:Y\in \mathcal{F}\}$.
If $X\in\mathcal{F}$ is a feasible set then the \emph{contraction} of $X$ yields the greedoid $G/ X=(S-X,\mathcal{F}/ X)$, where $\mathcal{F}/ X=\{Y\subseteq S-X:Y\cup X\in \mathcal{F}\}$.
A minor of $G$ is obtained by applying the two operations on $G$.

Equivalently, one can work in the language view:
Elements of the finite ground set (also called alphabet) $S$ are referred to as \emph{letters}.
The set of all finite sequences of $S$ is denoted as $S^*$.
Elements of $S^*$ are referred to as \emph{words}.
A set $\mathcal{L}\subseteq S^*$ is referred to as a \emph{language}.
An empty word is denoted by $\emptyset$.
The set of letters of a word $\alpha$ is denoted by $\tilde{\alpha}$, and the length of a word $\alpha$ is denoted by $|\alpha|$.
$\mathcal{L}$ is a \emph{simple language} if no letter of $S$ appears more than once in any word of $\mathcal{L}$. The concatenation of two words $\alpha,\beta\in S^*$ is simply denoted by $\alpha\beta$ and $\alpha$ is called a prefix of $\alpha\beta$.

\begin{definition}[Greedoid Language]\label{def:greedoid_lang}
	A greedoid language $G=(S,\mathcal{L})$ is a pair consisting of a finite ground set $S$ and a simple language $\mathcal{L}$ on $S$, if the following properties hold:
	\begin{enumerate}
		\item $\emptyset\in \mathcal{L}$
		\item $\alpha\beta\in \mathcal{L}$ implies $\alpha\in \mathcal{L}$
		\item\label{def:greedoid_lang:3} If $\alpha,\beta \in \mathcal{L}$ and $|\alpha|>|\beta|$ then there exists an $x\in \tilde{\alpha}$ such that $\beta x \in\mathcal{L}$
	\end{enumerate}
\end{definition}
\noindent
A \emph{basic word} of a greedoid language $G=(S,\mathcal{L})$ is a word $\alpha\in \mathcal{L}$ of maximum length. All basic words in a greedoid language have the same length. The set of basic words is denoted by $\mathcal{B}$.

Let $G=(S,\mathcal{L})$ be a greedoid language and $w:\mathcal{L}\to\mathbb{R}$ an objective function. Assume that we are interested in finding a basic word $\beta\in\mathcal{B}$ that minimizes $w(\beta)$ across all basic words of $G$.
For every $\alpha\in\mathcal{L}$ the \emph{set of continuations} of $\alpha$ is defined as $\Gamma(\alpha)=\{x\in S-\tilde{\alpha}:\alpha x \in \mathcal{L}\}$. Then the \emph{greedoid greedy algorithm} for the above problem can be described as follows \citep{korte_greedoids_1991}:

\begin{enumerate}
	\item Set $\alpha=\emptyset$
	\item If $\Gamma(\alpha)=\emptyset$ then stop and output $\alpha$
	\item Choose an $x\in\Gamma(\alpha)$ such that $w(\alpha x)\leq w(\alpha y)$ for every $y\in \Gamma(\alpha)$
	\item Replace $\alpha$ by $\alpha x$ and repeat at Step 2.
\end{enumerate}

\begin{theorem}[\cite{szeszler_sufficient_2022} Theorem 6]\label{thm:greedoid}
	Let $G=(S,\mathcal{L})$ be a greedoid language and $w:\mathcal{L}\to \mathbb{R}$ an objective function. Assume that the following condition holds:
	\begin{enumerate}
		\item \label{thm:greedoid:cond} If $\alpha x\in\mathcal{L}$ such that $w(\alpha x)\leq w(\alpha y)$ for every $y\in\Gamma(\alpha)$ and $\gamma=\alpha z\beta\in\mathcal{B}$ is a basic word then there exists a basic word $\delta=\alpha x \epsilon \in \mathcal{B}$ such that $w(\delta)\leq w(\gamma)$.
	\end{enumerate}
	Then the greedoid greedy algorithm finds a basic word of minimum weight with respect to $w$.
\end{theorem}

\begin{corollary}[\cite{szeszler_sufficient_2022} Corollary 1]
	\label{corollary:greedoid}
	If \cref{thm:greedoid:cond} in \cref{thm:greedoid} is violated in the greedoid language $G=(S,\mathcal{L})$ with respect to the objective function $w:\mathcal{L}\to\mathbb{R}$ then there exists an $\alpha\in\mathcal{L}$ such that a legal running of the greedy algorithm in the minor $G/\tilde{\alpha}$ gives a basic word that is not of minimum weight with respect to the objective function $\bar{w}(\beta)=w(\alpha \beta)$.
\end{corollary}

In other words, if \cref{thm:greedoid:cond} in \cref{thm:greedoid} is violated, then the greedy algorithm might give a suboptimal base in the minor of the greedoid.

\subsubsection{Proof of optimality}
\cref{algo:source} is a greedy algorithm operating on the greedoid language $G=(\mathcal{S}, \mathcal{L})$ where $\mathcal{L}$ is the set of all sampling orders on a vine of dimension $d$ on the set $\mathcal{S}=\{0,\dots,d-1\}$.
The algorithm constructs a basic word of this greedoid language, by extending a sampling order with a new element as guided by the objective function $\Call{Query}$, which we compute via the oracle subroutine and the algorithm seeks to minimize.

It is easy to see that the first two in \cref{def:greedoid_lang} hold.
The \cref{def:greedoid_lang:3} in \cref{def:greedoid_lang} holds by the recursive structure of a vine \citep{czado_analyzing_2019}. For a sampling order $\alpha$ with length $|\alpha|=a$ on a vine of dimension $d\geq a$, by dropping the first $d-a$ levels in the vine (\cref{def:vine_mst}), the remaining levels altogether are still a vine of dimension $d$ on the set $\mathcal{S}_{\alpha}=\{i|i\in\alpha\}$ on which $\alpha$ is a valid sampling order.
Since $\alpha$ is of length $a$, it must have covered all elements in $\mathcal{S}_{\alpha}$. For any other sampling order $\beta$ with $|\beta|<a$ there exists $x\in\tilde{\alpha}$ such that $\beta x$ is a valid sampling order.

With \cref{corollary:greedoid}, if \cref{algo:source} gives a non-optimal sampling order $\alpha$ on $G=(\mathcal{S}, \mathcal{L})$ and a vine on $\mathcal{S}$, then the algorithm in the minor $G/\tilde{\alpha}$ gives a basic word that is not of minimum weight. From the definition of vine, we know that the minor $G/\tilde{\alpha}$ corresponds to finding an optimal sampling order for the subvine on $\mathcal{S}\setminus\tilde{\alpha}$ whose dimension smaller than $d$ \citep{cooke_sampling_2015,zhu_common_2020,zhu_simplified_2021}.
It is easy to verify that the algorithm gives optimal sampling order on all vines of dimension $2$ and $3$, and thus the \cref{thm:greedoid:cond} in \cref{thm:greedoid} holds for vines of finite dimension $d$.
Therefore, \cref{algo:source} produces a sampling order whose number of h-function calls is minimal among all sampling orders on a given vine of finite dimension $d$.

\section{Experiments details} \label{sec:expriment_details}
% All headings should be lower case (except for first word and proper nouns), flush left, and bold.
\subsection{Compute resources}
% \TYcom{Comment: checklist point 8; cpu/ram/gpu/os/python/cost/more\\}
All experiments were run on a single workstation under \texttt{WSL2-x86/64 (Linux-5.15.167.4, glibc 2.35)} with an AMD Ryzen 9 5900X (12 cores, 64GB RAM) and an NVIDIA GeForce RTX 3060 GPU (12GB VRAM).
Software versions were \texttt{Python 3.13.2 (Clang 20.1.0)}, \texttt{PyTorch 2.7.0+cu126}.

\begin{itemize}
	\item Benchmarking (pyvinecopulib vs. torchvinecopulib): Fit, sampling, and density tasks across dimensions $d{=}10{-}50$ and sample sizes $n{=}1{,}000{-}50{,}000$, each repeated for $100$ times over the grid of test configurations, consumed a total of roughly $200$ hours.
	\item VCAE experiments ($100$ random seeds on MNIST): required roughly $1$ GPU hour.
	\item Prediction interval experiments ($100$ random seeds on California Housing and Online News Popularity): required roughly $190$ GPU hours in aggregate.
	\item Preliminary studies (hyperparameter sweeps, ablations): consumed roughly additional $300$ GPU hours, which are not included in the reported timings.
\end{itemize}

\subsection{Experimental setting/details}
% \TYcom{Comment: checklist point 6\\}

In all experiments, for vines, marginal distributions were estimated following \cite{o2014reducing,o2016fast}, and bivariate copulas were fit via the non-parametric transformation local likelihood kernels\citep{nagler2017nonparametric}, truncated with Kendall's $\tau$ independence test threshold fixed at $0.01$.

\subsection{VCAE}
We use a fully connected autoencoder (AE) architecture without convolutions. The encoder maps MNIST images of size $28 \times 28$ to a 10-dimensional latent space via the following layers: a flattening operation, a linear layer with 64 hidden units, followed by ReLU activation, another linear layer with 32 hidden units and ReLU activation, and a final linear layer projecting to $\mathbb{R}^{10}$. The decoder is symmetric: a linear layer with 32 units and ReLU activation, followed by a linear layer with 64 units and ReLU activation, and a final linear layer projecting back to the input dimension ($784$), with a sigmoid activation to ensure outputs lie in $[0,1]$. All models are trained using the Adam optimizer with a learning rate of $2 \times 10^{-4}$ and mean squared error (MSE) as the reconstruction loss. When refitting the model after training the vine, we add the negative log-likelihood of the latent code under a learned vine copula prior to the loss. Models are trained for 10 epochs using a batch size of 512 (on GPU) or 64 (on CPU), with the float32 matrix multiplication precision set to \texttt{medium} for better performance on modern CUDA devices.

\subsection{Prediction intervals}
For each dataset tested in \cref{sec:pred_intvl}, we use repeated random $70\%/10\%$ train/validation split after a $20\%$ test set held fixed.

For the base encoder-regressor (used for MC-dropout and deep ensembles): we use the MLP mapping the $d$-dimensional input to a $L$-dimensional latent code via successive linear layers of sizes
$d\to128\to64\to L\to32\to16\to8\to L$,
each followed by BatchNorm, LeakyReLU (0.1) and dropout $p$, and then a two-layer ''head'' $L\to16\to1$ (LeakyReLU and dropout $p/2$ between).
For deep ensembles, we train $M$ independent copies of the above encoder-regressor, each seeded differently, with dropout disabled at inference. At test time we take the empirical quantiles of their $M$ predictions for our quantiles and PIs.
For MC-dropout, we use the same base model but leave all dropout layers in train model at inference and draw $T$ forward passes, then compute the $\alpha/2$ and $1-\alpha/2$ quantiles across those $T$ samples.
For Bayesian NN, we replace every \texttt{nn.Linear} in the base encoder-regressor (including both ``encoder" and ``head") with \texttt{BayesianLinear} from \texttt{blitz} and adopt variational estimator \citep{esposito2020blitzbdl}. We train by minimizing the usual ELBO, and at test time draw $T$ posterior samples to form our PIs.

All of the above architectures were trained with Adam \citep{kingma2014adam} for up to $150$ epochs (early-stopped on validation MSE).
We selected hyperparameters: learning rate $\text{lr}\in \{1\mathrm{e}-4,1\mathrm{e}-3,1\mathrm{e}-2\}$, weight decay $\beta\in\{1, 0.91, 0.99\}$, dropout (where applicable) $p\in\{0.1,0.3,0.5\}$, via a grid search on a $5$-fold cross-validation on the combined training/validation pool, and the held-out test set was never accessed during tuning or training. The chosen values were $\text{lr}=1\mathrm{e}-3,~\beta=0.91,~p=0.3$.
Resulting metrics are collected over $100$ random train/validation set splits and initializations.
\section{Broader impacts} \label{sec:BroaderImpacts}
% All headings should be lower case (except for first word and proper nouns), flush left, and bold.
While this work is primarily methodological, we have reflected on its broader ethical, societal, and environmental implications as follows:

\paragraph{Positive impacts}
\begin{itemize}
	\item \textbf{More trustworthy AI:} By enabling scalable, flexible modeling of complex dependencies and uncertainty quantification,the VCG can improve the reliability of ML systems in safety-critical domains (e.g., autonomous vehicles, healthcare diagnostics) and in high-stakes decision-making in finance or scientific research.
	\item \textbf{Open, auditable tooling:} We release \texttt{torchvinecopulib} under an MIT license alongside full documentation, example notebooks, and explicit dependency licenses, supporting transparency, reproducibility, and community scrutiny.
	\item \textbf{Efficiency gains:} GPU vectorization and memory-efficient graph traversal reduce wall-clock time and peak memory, which can lower the per-run energy footprint compared to less-optimized implementations.
\end{itemize}

\paragraph{Potential negative impacts \& Misuse}
\begin{itemize}
	\item \textbf{Fairness and discrimination:} More powerful dependency models could be used in automated decision systems (e.g., credit scoring, hiring) without adequate audit for bias, potentially exacerbating unfair outcomes against protected groups.
	\item \textbf{Privacy and surveillance:} Downstream users might apply the VCG to sensitive or proprietary datasets (e.g., location traces, medical records). Without privacy safeguards, enhanced modeling could facilitate invasive profiling or tracking.
	\item \textbf{Financial misuse:} In quantitative finance, faster, more accurate risk models might be repurposed for speculative or predatory trading strategies, potentially destabilizing markets.
	\item \textbf{Environmental cost:} Large-scale or frequent retraining of high-dimensional models can still incur substantial energy use and associated carbon emissions.
\end{itemize}

\paragraph{Impact mitigation measures}
To address these risks, we encourage further community best practices:
\begin{itemize}
	\item \textbf{Encouraging privacy-preserving use:} We recommend users apply standard anonymization or differential-privacy techniques when fitting vines to sensitive datasets.
	\item \textbf{Fairness auditing:} We urge practitioners to evaluate downstream models for disparate impact across demographic groups, and to incorporate bias-mitigation strategies (e.g., reweighting, adversarial debiasing) when appropriate.
	\item \textbf{Energy transparency:} Our benchmarking scripts report both runtime and empirical quantiles over repeated runs, enabling users to estimate energy consumption; we plan to extend this to include direct measurements (e.g., via NVIDIA's NVML) in future releases.
\end{itemize}

Overall, the direct societal impact depends on downstream choices.
We provide the tools, documentation, and licensing needed for responsible use, and we encourage the community to pair our methods with established privacy, fairness, and sustainability practices.

\end{document}